%% file: main.tex
\documentclass[10pt]{article}

\usepackage[a4paper, margin=2.5cm]{geometry}
\usepackage[utf8]{inputenc}
\usepackage[T1]{fontenc}
\usepackage{inconsolata}
\usepackage[scaled=.90]{helvet}
\usepackage{microtype}
\usepackage{url}
\usepackage{graphicx}
\usepackage{subcaption}
\usepackage{booktabs}
\usepackage{multirow}
\usepackage[dvipsnames,table]{xcolor}
\PassOptionsToPackage{hypertexnames=false}{hyperref}
\usepackage[colorlinks]{hyperref}
\hypersetup{
    citecolor=[RGB]{50,100,170},
    linkcolor=[RGB]{50,100,170},
    urlcolor=[RGB]{255,102,178}}
\usepackage{fancyhdr}
\usepackage{natbib}
\bibpunct{(}{)}{;}{a}{,}{,}

\usepackage{amsmath}
\usepackage{amssymb}
\usepackage{mathtools}
\usepackage{amsthm}
\usepackage[capitalize,noabbrev]{cleveref}

\usepackage{caption}
\usepackage{array}
\usepackage{adjustbox}
\usepackage{placeins}
\usepackage{float}
\usepackage{wrapfig}

\newcommand{\toptitlebar}{%
  {\color{black}\hrule height 1pt}%
  \vskip 0.25in%
}
\makeatletter
\renewcommand{\maketitle}{%
  \begin{center}%
    \toptitlebar
    \vskip 0.1in%
    {\LARGE\bfseries \@title \par}%
    \vskip 0.3in%
    {\normalsize \@author \par}%
  \end{center}%
  \par
  \vskip 0.3in%
}
\renewcommand\section{\@startsection {section}{1}{\z@}{-2.0ex plus
    -0.5ex minus -.2ex}{1.5ex plus 0.3ex minus .2ex}{\large\bfseries\raggedright}}
\renewcommand\subsection{\@startsection{subsection}{2}{\z@}{-1.8ex plus
    -0.5ex minus -.2ex}{0.8ex plus .2ex}{\normalsize\bfseries\raggedright}}
\renewcommand\subsubsection{\@startsection{subsubsection}{3}{\z@}{-1.5ex plus
   -0.5ex minus -.2ex}{0.5ex plus .2ex}{\normalsize\bfseries\raggedright}}
\renewenvironment{abstract}%
  {\centerline{\large\bfseries Abstract}%
   \begin{list}{}%
      {\setlength{\rightmargin}{0.6cm}%
       \setlength{\leftmargin}{0.6cm}}%
    \item[]\ignorespaces}%
  {\unskip\end{list}}
\makeatother

\theoremstyle{plain}

\theoremstyle{definition}

\theoremstyle{remark}

\newcounter{obsCounter}

\usepackage{tcolorbox}
\tcbuselibrary{skins}
\newtcolorbox[auto counter]{observation}{
  colback=blue!5!white,
  colframe=black,
  colbacktitle=blue!80!white,
  coltitle=white,
  fonttitle=\bfseries,
  title=Observation~\thetcbcounter
}
\definecolor{takeawaybackground}{HTML}{EFF5F8}
\newtcolorbox[auto counter]{takeaway}[1][Takeaway~\thetcbcounter]{
  enhanced,
  colback=takeawaybackground,
  colframe=black,
  colbacktitle=black,
  coltitle=white,
  fonttitle=\bfseries,
  title={#1},
  attach boxed title to top left={xshift=4mm,yshift=-\tcboxedtitleheight/2},
  boxed title style={boxrule=0pt,arc=1pt,left=2mm,right=2mm,top=.5mm,bottom=.5mm},
  top=4mm,
  before skip=10pt,after skip=6pt
}
\newcommand{\D}{\mathcal D}
\newcommand{\E}{\mathbb E}
\newcommand{\SFT}{\mathrm{SFT}}

\title{Reinforcement Learning Can Be Recovered \\
From Its Own Rollouts}
\author{
  \textbf{Jiayi Sheng}$^{*1}$ \quad
  \textbf{Jingwen Gu}$^{*2}$ \quad
  \textbf{Guanning Zeng}$^{1}$ \quad
  \textbf{Zhaoyi Zhou}$^{1}$ \\
  \textbf{Stefano Viel}$^{3}$ \quad
  \textbf{Qianqian Wang}$^{4}$ \quad
  \textbf{Fahim Tajwar}$^{1}$ \quad
  \textbf{Andrea Zanette}$^{1}$ \\
  \vspace{1mm}
  $^1$Carnegie Mellon University \quad $^2$University of Illinois Urbana-Champaign \\
  $^3$Tufa Labs \quad $^4$Harvard University \\
  \vspace{1mm}
  \texttt{\{jsheng2,azanette\}@andrew.cmu.edu}, \texttt{jgu17@illinois.edu} \\
}

\begin{document}

\maketitle
\renewcommand{\thefootnote}{}
\footnotetext{$^*$Equal contribution.}
\renewcommand{\thefootnote}{\arabic{footnote}}
\thispagestyle{fancy}
\fancyhead{}
\lhead{\raisebox{-0.7cm}{\includegraphics[height=0.4cm]{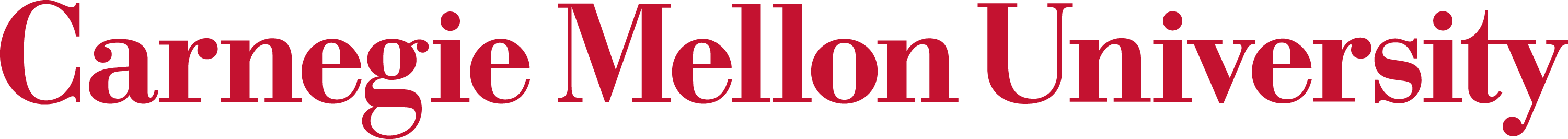}}}
\renewcommand{\headrulewidth}{0pt}
\setlength{\headheight}{18pt}
\setlength{\headsep}{3mm}
\suppressfloats[t]

\input{content/0_Abstract}

\input{content/1_Introduction}
\input{content/2_Preliminaries}

\input{content/3_Recovery}
\input{content/4_Applications}
\input{content/5_Related_Work}
\input{content/6_Conclusion}

\section*{Acknowledgements}
This research was supported in part by Impossible Inc., Jane Street, and Amazon AGI.

\bibliography{references}
\bibliographystyle{plainnat}

\newpage
\appendix
\raggedbottom

\section{Experimental Settings}\label{sec:app_settings}
All online runs and replays are implemented in the verl framework. Unless noted otherwise, maximum likelihood replay minimizes the token-level negative log-likelihood of \emph{all} archived responses, correct and incorrect, starting from the same initialization as the online run, with one update per archived RL step in the original order. Where the MaxRL advantage is used, a response with reward $r$ has advantage $(r-\hat p)/(\hat p+10^{-6})$, where $\hat p$ is the mean reward of the rollouts sampled for the same prompt (or image).

\paragraph{Replay updates.}
Each archived batch $D_t$ contains $B$ prompts $x_i$ with $G$ responses $y_{ij}$ per prompt.
Let $y_{ij,l}$ denote token $l$ of response $y_{ij}$ and let $N_t=\sum_{i,j}|y_{ij}|$ be the number of response tokens in the batch. The empirical SFT loss is
\begin{equation}\label{eq:sft}
\mathcal L_{\SFT}(\hat\pi_t;D_t)=-\frac{1}{N_t}\sum_{i,j}\sum_{l=1}^{|y_{ij}|}\log\hat\pi_t(y_{ij,l}\mid x_i,y_{ij,<l}).
\end{equation}

For importance-sampled RL replay, let $r_{ij}=r(x_i,y_{ij})$, $b_{ij}$ be the reward baseline for this response, and $w_i$ the prompt weight. Write $\pi(y_{ij,l})$ for $\pi(y_{ij,l}\mid x_i,y_{ij,<l})$.
We truncate the ratio between the replayed policy and the rollout-generating policy at $C=2$,
\begin{equation}\label{eq:is_weight}
\rho_{ij,l}=\min\!\big(\hat\pi_t(y_{ij,l})/\pi_t(y_{ij,l}),\,C\big),
\end{equation}
and use the update
\begin{equation}\label{eq:is_objective}
\hat g^{\mathrm{IS}}_t=\frac{1}{BG}\sum_{i,j}w_i\sum_{l=1}^{|y_{ij}|}\rho_{ij,l}\,\nabla\log\hat\pi_t(y_{ij,l})\,\big(r_{ij}-b_{ij}\big).
\end{equation}

\paragraph{Recovery metrics.}
Let $F_k(\pi)$ denote the pass@$k$ of policy $\pi$ on a given benchmark.
We average the absolute gap between the replayed and online policies over $\mathcal K=\{1,2,4,\ldots,32\}$, omitting $k=2$ for Qwen3-4B-Base.
The trajectory error also averages over the set $\mathcal T$ of evaluation steps,
\begin{equation}\label{eq:traj}
\epsilon_{\mathrm{traj}}=\frac{1}{|\mathcal T|\,|\mathcal K|}\sum_{t\in\mathcal T}\sum_{k\in\mathcal K}\big|F_k(\hat\pi_t)-F_k(\pi_t)\big|,
\end{equation}
and the endpoint error compares the two final checkpoints,
\begin{equation}\label{eq:end}
\epsilon_{\mathrm{end}}=\frac{1}{|\mathcal K|}\sum_{k\in\mathcal K}\big|F_k(\hat\pi_T)-F_k(\pi_T)\big|.
\end{equation}

\subsection{Language-Model Settings}\label{sec:app_lm}
Table~\ref{tab:app_lm} lists the settings behind Figures~\ref{fig:replay}, \ref{fig:replay_end}, and~\ref{fig:onebatch}. Polaris models are evaluated on MATH-500, AIME25, GPQA-Diamond, Minerva \citep{lewkowycz2022minerva}, and BeyondAIME~\citep{seed2025thinking}. All language-model evaluations sample at temperature 0.6 and top-$p$ 0.95, and report unbiased pass@$k$ unless noted otherwise.

For maximum likelihood replay on the last $m$ steps (Section~\ref{sec:recovery_few_rollouts}), the student loops over the batches $D_{T-m},\ldots,D_{T-1}$ in order, one update per batch, and starts again from $D_{T-m}$ after $D_{T-1}$. It stops after the number of updates in Table~\ref{tab:app_lm}.

\begin{table}[H]
\centering
\small
\caption{Language-model settings. Replay denotes maximum likelihood replay unless an offline RL algorithm is specified. A ``step'' is one online RL update; one replay update consumes the rollouts of one archived step.}
\label{tab:app_lm}
\begin{tabular}{p{0.16\linewidth}p{0.24\linewidth}p{0.28\linewidth}p{0.21\linewidth}}
\toprule
Setting & Online MaxRL & Replay & Evaluation \\
\midrule
Qwen3-1.7B-Base, Polaris-53K & 1000 steps, 256 prompts $\times$ 16 rollouts $=4096$ per step. lr $10^{-6}$, constant & Replay: AdamW, lr $10^{-6}$, constant, wd 0.01, clip 1.0, max length 6144; 1000 updates. Last step only: the 4096 rollouts of step 1000, 200 updates. Offline MaxRL: same rollouts and lr, with the online MaxRL advantage $(r-\hat p)/(\hat p+10^{-6})$ computed from the 16 stored rewards per prompt, one update per stored step & Figures~\ref{fig:replay} and~\ref{fig:replay_end}: 128 samples per question, every 100 steps. Figure~\ref{fig:onebatch}: 512 samples \\
\addlinespace
Qwen3-4B-Base, Polaris-53K & 1000 steps, 256 prompts $\times$ 16 rollouts $=4096$ per step; lr $10^{-6}$, constant & Replay on the archived rollouts in order: AdamW, lr $10^{-6}$, constant, wd 0.01, clip 1.0, max length 6144; 1000 updates & Figure~\ref{fig:replay_end}: 1024 samples per question, final checkpoint \\
\addlinespace
Qwen3-1.7B-Base, Knights and Knaves & 1000 steps, 256 puzzles $\times$ 16 rollouts per step. lr $10^{-6}$, constant & Replay: AdamW, lr $10^{-6}$, constant, wd 0.01, max length 8192; 1000 updates. Last step only: the rollouts of step 1000, evaluated after 500 updates & Replay: 32 samples per puzzle. Online: 32 samples per puzzle every 100 steps, 128 at the final checkpoint \\
\addlinespace
SmolLM2-360M-Instruct, GSM8K & 1500 steps, 256 prompts $\times$ 128 rollouts per step; AdamW, lr $10^{-5}$, constant; response length 2048 & Replay: AdamW, lr $10^{-5}$, constant, on the last 1, 5, or 50 steps; all compared after 420 updates & 32 samples per question; pass@$k$ for $k>1$ is the bootstrap estimate computed by verl \\
\bottomrule
\end{tabular}
\end{table}

\subsection{Maze}\label{sec:app_maze}
The policy is trained to solve the $17\times17$ maze task of \citet{tajwar2026maxrl}. The online MaxRL run and all replays start from a warm-started checkpoint after SFT on demonstrations for 1500 steps.
Table~\ref{tab:app_maze} lists the settings for Figure~\ref{fig:different_tasks} (left). The all-rollouts run replays each of the 1000 RL steps 8 times, and a last-$k$ run replays each of the last $k$ steps $8000/k$ times, so both take about 8000 replay updates. For $k>1$, pass@$k$ is the bootstrap estimate computed by verl from the evaluation samples.

\begin{table}[H]
\centering
\small
\caption{Maze settings.}
\label{tab:app_maze}
\begin{tabular}{p{0.22\linewidth}p{0.33\linewidth}p{0.37\linewidth}}
\toprule
 & Online MaxRL & Maximum likelihood replay \\
\midrule
Initialization & \texttt{maze\_ckpt1500} & same \\
Data per step & 256 mazes $\times$ 32 rollouts $=8192$ & one archived RL step (8192 rollouts) \\
Length (prompt / response) & 320 / 180 tokens & 512 tokens in total \\
Objective & MaxRL advantage, no KL & NLL on all rollouts \\
Optimizer & AdamW, lr $10^{-4}$, constant & AdamW, lr $3\times10^{-5}$, constant, wd 0.01, clip 1.0 \\
Rollout sampling & temperature 1.0 & -- \\
Replayed steps & -- & all: steps 1--1000, $8\times$ each; last $k$: steps $1001-k$ to 1000, $8000/k\times$ each \\
Evaluation & RL step 1000, 2048 samples per maze & last checkpoint, 32 samples per maze \\
Eval sampling & temperature 1.0, top-$p$ 1.0 & same \\
\bottomrule
\end{tabular}
\end{table}

\subsection{ImageNet}\label{sec:app_imagenet}
We train ResNet-50~\citep{he2016resnet} on ImageNet~\citep{deng2009imagenet} with online MaxRL and compare maximum likelihood replay, offline REINFORCE, and offline MaxRL on its archived rollouts (Figure~\ref{fig:different_tasks}, right). Table~\ref{tab:app_imagenet} lists the settings.
\begin{table}[H]
\centering
\small
\caption{ImageNet settings.}
\label{tab:app_imagenet}
\begin{tabular}{p{0.22\linewidth}p{0.33\linewidth}p{0.37\linewidth}}
\toprule
 & Online MaxRL & Offline replay \\
\midrule
Initialization & ResNet-50  & same \\
Data per step & 256 images $\times$ 1024 rollouts $=262144$ & one archived RL step (262144 rollouts) \\
Input resolution & $224\times224$ (RandomResizedCrop from $256\times256$) & same ($224\times224$ center crop for eval) \\
Objective & MaxRL advantage, no KL & maximum likelihood replay: NLL on all rollouts; Offline REINFORCE: NLL on correct rollouts only; Offline MaxRL: all rollouts, weighted by the MaxRL advantage $(r-\hat{p})/(\hat{p}+10^{-6})$ \\
Optimizer & SGD, lr $0.1$, momentum $0.9$, wd $10^{-4}$, cosine decay with 2-epoch warm-up, clip $10^{5}$ & same \\
Rollout sampling & softmax temperature 1.0, multinomial with replacement, $K=1024$ per image & -- \\
Replayed steps & -- & all: epochs 1--20, each online step's rollouts replayed once \\
Evaluation & every 1000 steps, analytic pass@$k$ for $k=1,2,4,\dots,1024$ on the ImageNet validation set & every 1000 steps and at the end of each replayed epoch, same metric \\
Eval sampling & -- (pass@$k$ computed analytically from the softmax probabilities, without resampling) & same \\
\bottomrule
\end{tabular}
\end{table}

\section{Additional Recovery Results}\label{sec:appendixreplay}

Table~\ref{tab:recovery_combined} summarizes the trajectory and endpoint recovery errors from Section~\ref{sec:rollout_recovery}. This section also covers rollouts of other online RL algorithms, KnK puzzles of each size, the number of last-step rollouts needed, and tasks beyond language modeling.

\input{tables/recovery_errors_combined}

\subsection{Other Online RL Algorithms}\label{sec:app_online_algs}

\begin{figure}[H]
\centering
\includegraphics[width=\textwidth]{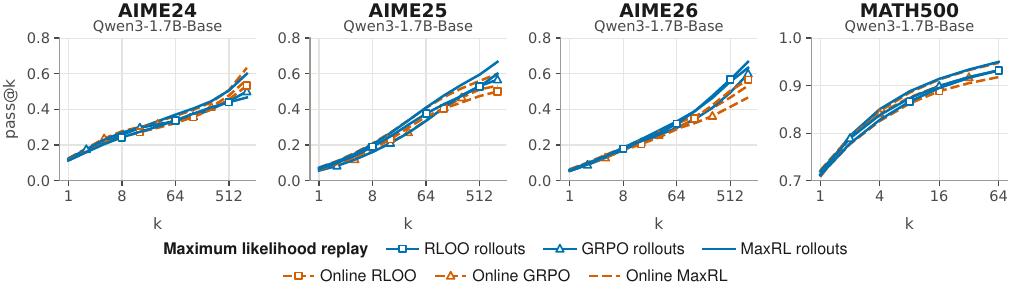}
\caption{\textbf{Chronological maximum likelihood replay recovers the final checkpoint of different online RL algorithms.} Final-checkpoint pass@$k$ of Qwen3-1.7B-Base fine-tuned on the archived rollouts of MaxRL, GRPO, and RLOO runs, each compared with the corresponding online run. Solid blue shows maximum likelihood replay and dashed vermilion shows online RL. Curves are unmarked for MaxRL, marked with triangles for GRPO, and marked with squares for RLOO.}
\label{fig:different_online_algs}
\end{figure}

In addition to MaxRL, we reconstruct the performance of online GRPO and RLOO by chronologically replaying the rollouts sampled from their training loops. We follow the same hyperparameters and experiment settings in Appendix~\ref{sec:app_lm} for both online RL and maximum likelihood replay, with the choice of online RL algorithm being the only difference. Figure~\ref{fig:different_online_algs} demonstrates that chronological replay also recovers online runs of GRPO and RLOO. In particular, for all three online RL algorithms, we observe that maximum likelihood replay surpasses the original online pass@$k$ for large $k$ on AIME25-26 and MATH500. Hence we conclude that recovery is a general phenomenon for various online RL algorithms rather than an artifact of the MaxRL algorithm.

\subsection{Knights and Knaves by Puzzle Size}\label{sec:app_knk}

\begin{figure}[H]
\centering
\includegraphics[width=\textwidth]{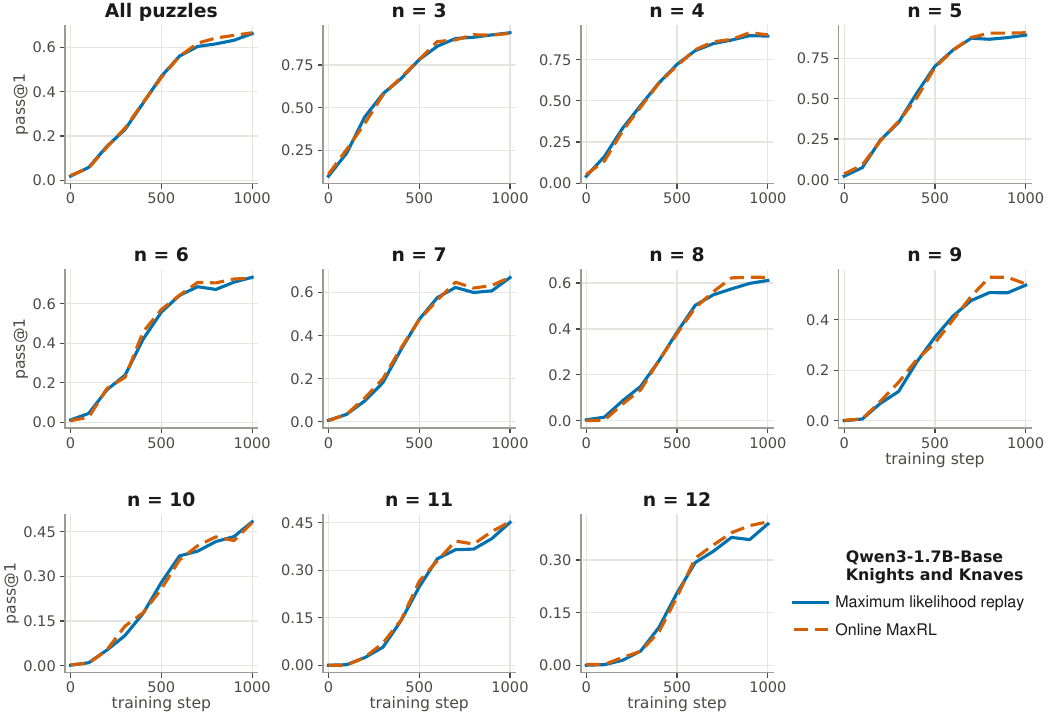}
\caption{\textbf{Knights-and-Knaves trajectory replay.} Pass@1 over training of online MaxRL (dashed vermilion) and of maximum likelihood replay on its archived rollouts in chronological order (blue), for all puzzles and for each puzzle size $n=3,\ldots,12$. Online MaxRL is evaluated with 4 samples per puzzle during training and 128 at the final checkpoint. We use 32 for replay. Small gaps appear late in training on some of the larger puzzles.}
\label{fig:knkapp}
\end{figure}

\begin{figure}[H]
\centering
\includegraphics[width=\textwidth]{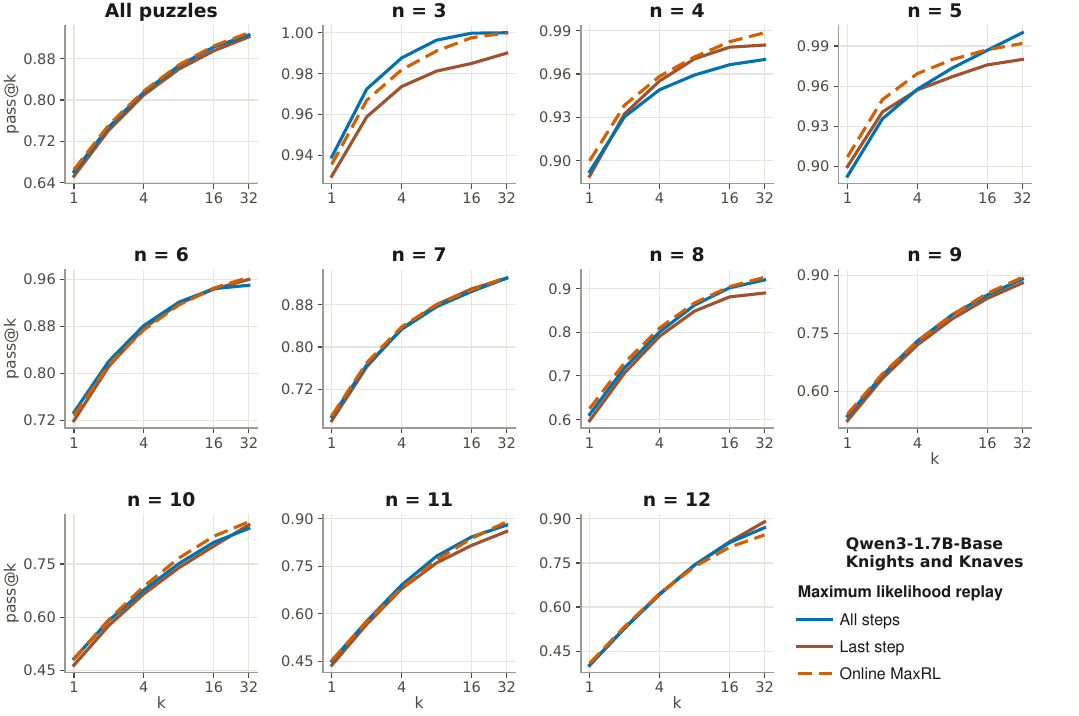}
\caption{\textbf{KnK endpoint reconstruction.} Final-checkpoint pass@$k$ of online MaxRL (dashed vermilion), of maximum likelihood replay on all 1000 archived steps (blue), and of replay on the last step alone (brown), for all puzzles and for each puzzle size $n$.}
\label{fig:knkendpoint}
\end{figure}

Figures~\ref{fig:knkapp} and~\ref{fig:knkendpoint} break down the KnK results of Section~\ref{sec:rollout_recovery} by puzzle size.

\subsection{How Many Rollouts of the Last Step Are Needed?}\label{sec:replay_amount}

For Qwen3-1.7B-Base, the 4096 rollouts of the last RL step recover the final checkpoint (Section~\ref{sec:recovery_few_rollouts}). To see how many of them are needed, we train the base model on random subsets of $n=16,32,\dots,4096$ rollouts. Each replay update uses the entire subset as a single batch, so all runs take the same 100 updates and differ only in the amount of data they see (Figure~\ref{fig:budget}).

\begin{figure}[H]
\centering
\includegraphics[width=\textwidth]{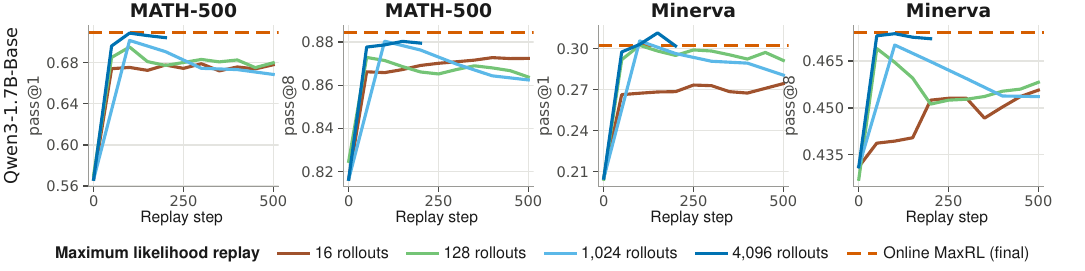}
\caption{\textbf{Maximum likelihood replay with different rollout budgets.} The runs of Figure~\ref{fig:budget} over replay updates, for 16, 128, 1024, and all 4096 of the last step's rollouts, compared with the final online MaxRL checkpoint (dashed vermilion); pass@8 is smoothed over 3 evaluations. The 4096-rollout run was trained for only 200 steps; its curve is not extrapolated beyond that point.}
\label{fig:budgetcurveapp}
\end{figure}

We compare each subset with maximum likelihood replay on all 4096 rollouts, which matches online MaxRL. On MATH-500, the pass@1 gap to the full batch shrinks steadily with $n$, from $0.033$ at 16 rollouts to $0.007$ at 1024, and closes only at 2048 rollouts ($0.708$ against $0.709$). Pass@8 is less demanding and matches the full batch from 1024 rollouts on. Minerva saturates much earlier: from 64 rollouts on, both pass@1 and pass@8 stay within $0.006$ of the full batch. Training longer does not make up for less data, and with additional updates some of the curves degrade (Figure~\ref{fig:budgetcurveapp}).

Matching the final policy thus takes most of the last step's rollouts on MATH-500, about 2048 of 4096, but far fewer on Minerva.

\subsection{Tasks Beyond Language Modeling}\label{sec:app_beyond}

\begin{figure}[H]
\centering
\includegraphics[width=\textwidth]{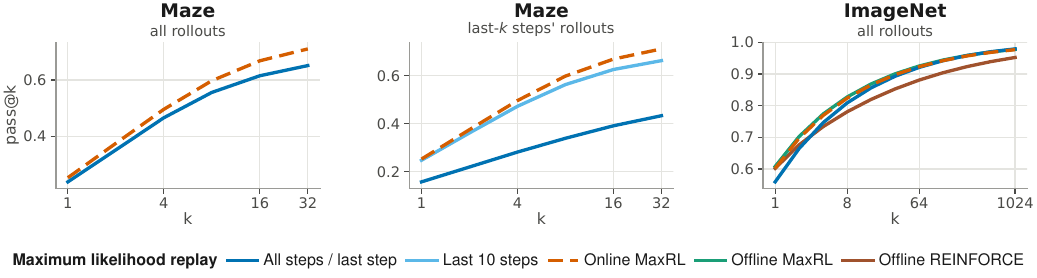}
\caption{\textbf{Reconstruction beyond language tasks.} Final-checkpoint pass@$k$ of online MaxRL (dashed vermilion) and of offline training on its archived rollouts, for Maze and ResNet-50 on ImageNet. The first Maze panel and the ImageNet panel use the rollouts of all RL steps, and the second Maze panel only those of the last $k$ steps.}
\label{fig:different_tasks}
\end{figure}

\begin{figure}[H]
\centering
\includegraphics[width=.66\textwidth]{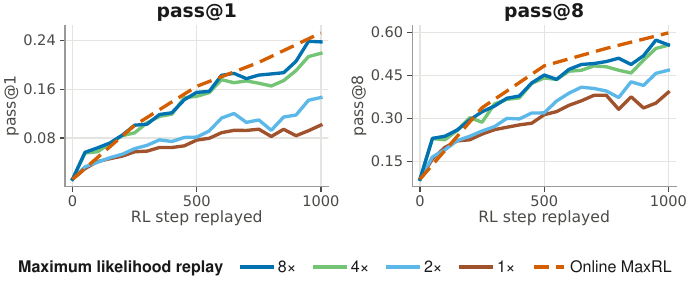}
\caption{\textbf{Maze trajectory reconstruction with repeated updates per saved RL step.} Maximum likelihood replay uses the rollouts of each of the 1000 online MaxRL steps 1, 2, 4, or 8 times before moving to the next step, compared with online MaxRL (dashed vermilion). This experiment varies the optimization intensity as well as the replay schedule, so it does not isolate the role of drift between adjacent policies.}
\label{fig:mazeupdates}
\end{figure}

A claim about RL assets should not rest on language-model reasoning alone. We therefore repeat the replay experiments on two tasks of a different nature: navigating a $17\times17$ maze, and image classification with a ResNet-50 \citep{he2016resnet} on ImageNet \citep{deng2009imagenet} (Figure~\ref{fig:different_tasks}). For each task we train offline on the rollouts of all steps of an online MaxRL run, on Maze also on those of its last $k$ steps only, and compare with the final online checkpoint. On Maze, maximum likelihood replay uses each step's rollouts 8 times before moving to the next step (Figure~\ref{fig:mazeupdates} compares 1, 2, 4, and 8 repetitions); a last-$k$ run repeats each of its $k$ steps $8000/k$ times, so that every Maze run takes about 8000 replay updates. Settings are in Appendix~\ref{sec:app_settings}.

\begin{observation}
Replay also recovers online MaxRL on Maze and ImageNet, and on Maze a short window of late rollouts is enough.
\end{observation}

On Maze, maximum likelihood replay on the rollouts of all 1000 RL steps reaches pass@1 of $0.238$ against $0.252$ online (Figure~\ref{fig:different_tasks}, left). Training on only the last step's rollouts falls short ($0.157$), but the last 10 steps already reach $0.247$ at the same replay budget.

On ImageNet, Offline MaxRL is within $0.006$ of online MaxRL for every $k\le 1024$ (Figure~\ref{fig:different_tasks}, right). Maximum likelihood replay starts lower (pass@1 $0.56$ against $0.60$) but is within $0.01$ from $k=16$ on, while Offline REINFORCE falls behind at larger $k$, by up to $0.046$.

\section{Additional Replay-Order Results}\label{sec:app_order}

\begin{figure}[H]
\centering
\includegraphics[width=\textwidth]{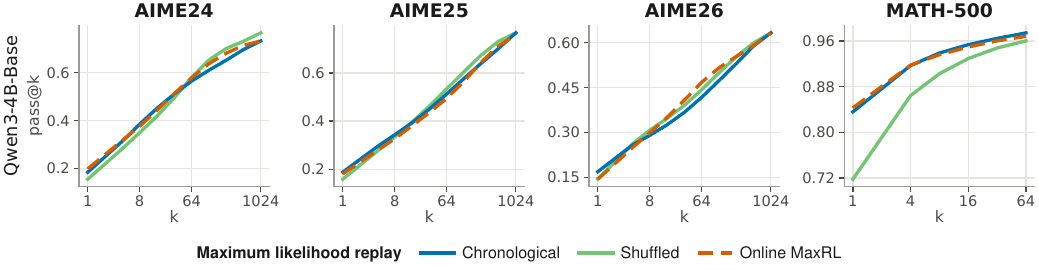}
\caption{\textbf{Shuffled-order maximum likelihood replay for Qwen3-4B on Polaris.} Final-checkpoint pass@$k$ of online MaxRL (dashed vermilion) and of maximum likelihood replay on its archived rollouts in chronological (blue) and shuffled (green) order.}
\label{fig:qwen3_4b_shuffled}
\end{figure}

\begin{figure}[H]
\centering
\includegraphics[width=\textwidth]{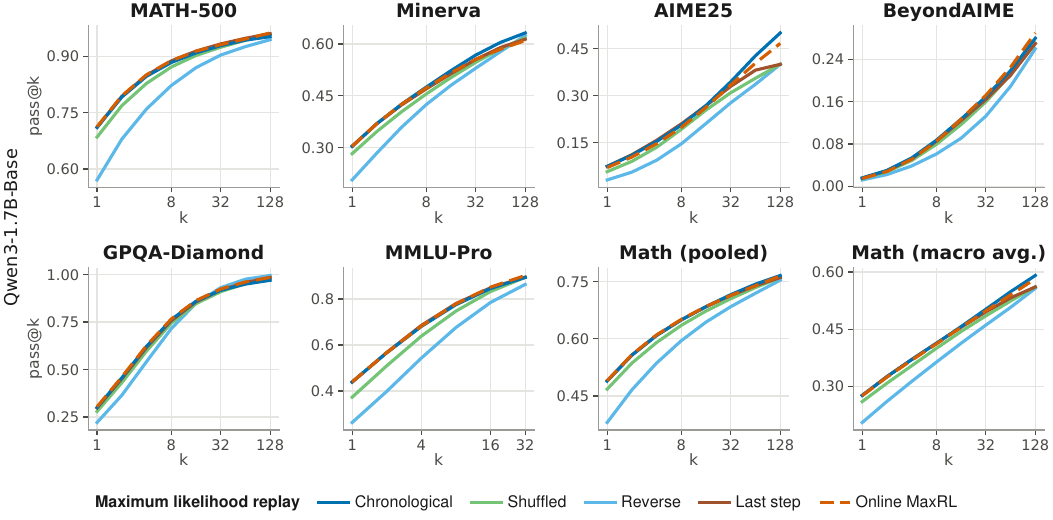}
\caption{\textbf{Out-of-domain evaluations.} Final-checkpoint (step 1000) pass@$k$ of the replay-order runs of Figure~\ref{fig:order} and of maximum likelihood replay on the last step's rollouts, including GPQA-Diamond and MMLU-Pro. Most benchmarks are evaluated with 128 samples per question; MMLU-Pro is evaluated on a 10\% subset with 32 samples per question, so its curves end at $k=32$.}
\label{fig:oodapp}
\end{figure}

Figure~\ref{fig:qwen3_4b_shuffled} repeats the shuffled-order experiment of Section~\ref{sec:recovery_order} for Qwen3-4B-Base. Figure~\ref{fig:oodapp} reports the final pass@$k$ of the replay-order runs on additional benchmarks, including MMLU-Pro~\citep{wang2024mmlupro}.

\section{Composition Details}\label{sec:app_composition}
\subsection{Composition in Data Space}\label{sec:compose_design}
Given the archives $\D^1,\ldots,\D^M$ of $M$ specialists $\pi^1,\ldots,\pi^M$ trained from the same base model $\pi_0$, data-space composition trains one student on all of them,
\begin{equation}
\pi_0\xrightarrow{\ \text{maximum likelihood replay on }\D^1,\ldots,\D^M\ }\hat\pi^{\mathrm{comp}}.
\label{eq:compose}
\end{equation}
Because the composed object is data, composition makes explicit choices that weight merging leaves implicit. We decompose it into three components.

\paragraph{Selection.} For each archive $\D^m=(D^m_0,\ldots,D^m_{T_m-1})$ of a run with $T_m$ steps, a subset of steps $\mathcal S_m\subseteq\{0,\ldots,T_m-1\}$ is chosen. Natural choices are the full trajectory, the last $K$ steps, or a single late-stage batch. Section~\ref{sec:recovery_few_rollouts} indicates that the right choice is task dependent: one late batch suffices for endpoint recovery on Polaris, whereas SmolLM2-360M requires a window of recent batches.

\paragraph{Scheduling.} The selected data are presented to the student either by \emph{joint-batch replay}, where every optimizer update contains samples from all tasks, or by \emph{alternating replay}, where consecutive updates are assigned to different tasks in a round-robin order. When full trajectories are used, batches can also be paired by RL step, so that the student follows all source trajectories in parallel.

\paragraph{Allocation.} Each task receives a share $w_m$ of the training signal, with $\sum_m w_m=1$. Task proportions must be measured in \emph{tokens}, not in numbers of responses or updates: under the token-normalized loss in Eq.~\eqref{eq:sft}, a task with longer responses receives more gradient weight per sample. Equal response counts can therefore produce a strongly unbalanced composition when response lengths differ across tasks.

\subsection{Experimental Protocol and Baselines}\label{sec:compose_protocol}
All specialists are trained with online RL from the same base model, and their rollouts are archived in chronological order as described in Section~\ref{sec:assets}. Unless stated otherwise, specialists are Qwen3-1.7B-Base models trained with MaxRL. The tasks are chosen to differ in domain, response format, and response length: mathematics (Polaris), logical puzzles (Knights and Knaves, KnK), and code (PIE~\citep{shypula2024pie}). We study the Polaris and KnK pair.

We compare our \emph{data-space composition} against three  \emph{weight-space} baselines: weight averaging, task arithmetic, and TIES-Merging \citep{yadav2023ties}.
Table~\ref{table:composition_math_knk} reports the full results of the composition experiment in Section~\ref{sec:rollout_specialists}.

\input{tables/math_knk_results}

We additionally ablate the allocation of training tokens across the math and KnK rollout pools. In the archived late-stage data, KnK rollouts are roughly twice as long as Polaris rollouts, so mixing equal numbers of responses gives KnK substantially more gradient weight. Consistent with this, count-balanced mixing degrades mathematics performance, which motivates allocation by tokens. We compare token allocations $w_{\mathrm{Polaris}}:w_{\mathrm{KnK}}$ of 1:1 and 2:1, as well as dynamic allocations that adjust $w_m$ during training. For alternating replay with last-step rollouts, last-25\% rollouts, and last-50\% rollouts, we report results with both 1:1 and 2:1 token allocations in Table~\ref{tab:math_knk_token_balanced}, with the latter replaying math training batches twice for each KnK training batch. We observe that training token allocation between the two rollout pools do not have a significant impact on recovery, and both allocation schemes recover the specialist pass@$k$ reliably.

\input{tables/math_knk_token_balanced}

\end{document}

%% file: content/0_Abstract.tex
\begin{abstract}
Broadly speaking, reinforcement learning (RL) post-training converts compute into model capabilities by training on model-generated data. Typically, the rollouts are discarded after each update, and the knowledge acquired during RL resides in the model weights. In this work, we investigate how much of this knowledge is retained in the rollouts themselves.
A natural first attempt is to replay the original RL algorithm on archived (rather than freshly generated) rollouts, but this generally leads to catastrophic collapse. We provide a simple explanation for this phenomenon, which in turn suggests a remedy we call maximum likelihood replay: fitting the model to its archived rollouts. Surprisingly, starting from the same base model and replaying the rollouts in the order in which they were generated, without any distinction between correct and incorrect responses, is sufficient to recover the performance of online RL. In fact, a single maximum likelihood update per rollout batch closely tracks the entire learning curve of online RL.
Furthermore, maximum likelihood replay on the rollouts of the last few RL steps often suffices to recover the final model's performance. This suggests that much of the compute invested in RL goes into discovering an improved sampling distribution rather than learning it from samples.

\end{abstract}
\vspace{-6mm}

%% file: content/1_Introduction.tex
\section{Introduction}
Reinforcement learning (RL) post-training converts compute into model capabilities by training on model-generated data \citep{lambert2025tulu3,guo2025deepseekr1}. Throughout training, the model repeatedly generates samples from its current distribution and uses reward feedback to improve this distribution. Generating these samples is particularly expensive for large language models due to their autoregressive nature \citep{noukhovitch2025asynchronous}. Yet, in standard on-policy post-training, the rollouts are typically discarded after the corresponding round of updates \citep{schulman2017ppo,shao2024deepseekmath,ahmadian2024back,tajwar2026maxrl}. At the end of training, what remains is the model weights, where the knowledge acquired during RL is encoded.

In this work, we investigate a different representation of this knowledge: the data generated by the model throughout training. As the model improves, so does the distribution from which its training data is sampled. This raises a natural question:
\begin{center}
\textbf{How much of what a model learns during reinforcement learning \\ can be recovered from the data it generates along the way?}
\end{center}

A natural first attempt is to reproduce the original RL run by replaying the archived rollouts, treating them as an offline RL dataset \citep{levine2020offline}. In principle, starting from the same base model and applying the same updates on the same data should reproduce the original training trajectory. In our experiments, however, this procedure is remarkably unstable and leads to catastrophic collapse, even with the off-policy corrections we evaluate \citep{degris2012offpolicy,marek2026score}. We identify a mechanism that can make replay unstable: once the replayed model deviates from the rollout-generating model, the policy-gradient baseline can amplify this discrepancy over successive updates.

This observation suggests a different approach. Rather than attempting to reproduce the original RL updates, we use \emph{maximum likelihood replay} to fit the model directly to samples from the distributions that generated the rollouts \citep{rusu2016policy,kim2016sequence}. At the population level, this objective pulls the replayed model toward the rollout-generating model, correcting rather than amplifying discrepancies between the two.

While this observation is intuitive at the population level, it is the empirical effectiveness of this approach that is surprising. Starting from the same base model and taking a single maximum likelihood update on each batch of rollouts, in the order in which they were generated, we recover with remarkable fidelity the entire learning curve of online RL. This is achieved without access to rewards or even distinguishing correct from incorrect responses. We observe close trajectory recovery across multiple settings, and recover final performance across different model sizes, tasks, and RL algorithms. Thus, the evolution of model performance throughout RL training can be recovered without reproducing the optimization process through which these capabilities were acquired.

Perhaps more surprisingly, recovering the final model's performance does not require replaying the entire training history. In several settings, repeated maximum likelihood updates on the rollouts from a single late-stage RL step recover the performance reached after a thousand online RL updates. This suggests that much of the computational effort in RL goes into discovering an improved sampling distribution rather than learning it from samples.

Beyond recovery, these archives support further reuse: rollouts from separately trained specialists can be combined to train one model, and rollouts generated by one model can train another.
Together, these findings show that archived RL rollouts can preserve both the progression and the final level of model performance, making them reusable assets of the training process.

%% file: content/2_Preliminaries.tex
\section{Preliminaries}\label{sec:assets}

\paragraph{The rollout archive.} At step $t$, the current policy $\pi_t$ receives a batch of prompts drawn from a distribution $\rho$ and samples several responses to each prompt. We denote the prompts and responses by $D_t$. A verifier assigns each response $y$ to prompt $x$ a binary reward $r(x,y)\in\{0,1\}$, and the online algorithm uses these rewards to update $\pi_t$ to $\pi_{t+1}$. We save the batches in order, giving the archive $\D=(D_0,\ldots,D_{T-1})$ for a run of $T$ steps (Figure~\ref{fig:method}). The last batch thus comes from $\pi_{T-1}$, just before the final update. Maximum likelihood replay uses only the prompts and responses. For RL replay we also save the rewards, and for importance sampling we record the probability of each sampled token given its prefix.

\paragraph{Online policy updates.} Fix a prompt $x$ and let $p_t(x)=\E_{y\sim\pi_t(\cdot\mid x)}[r(x,y)]$ be the probability of a correct response. At the population level, the algorithms in Table~\ref{tab:online_algs} use $p_t(x)$ as their reward baseline. With prompt weight $w_t(x)$, the contribution from $x$ to the update is
\begin{equation}\label{eq:pop_pg}
g_t(x)=w_t(x)\,\E_{y\sim\pi_t(\cdot\mid x)}
  \big[(r(x,y)-p_t(x))\,\nabla\log\pi_t(y\mid x)\big].
\end{equation}
\input{tables/online_algs}

\noindent Here $\nabla$ differentiates the log probability with respect to the parameters of the displayed policy. Since the on-policy score has mean zero, $\E_{y\sim\pi_t(\cdot\mid x)}[\nabla\log\pi_t(y\mid x)]=0$, the baseline term vanishes in expectation. The full update averages $g_t(x)$ over prompts drawn from $\rho$. Unit weights give the gradient of the mean success probability. Different choices of $w_t(x)$ lead to different objectives. Table~\ref{tab:online_algs} lists the weights for REINFORCE \citep{williams1992reinforce}, RLOO \citep{ahmadian2024back}, GRPO \citep{shao2024deepseekmath}, and MaxRL \citep{tajwar2026maxrl}.

%% file: tables/online_algs.tex
\begin{wraptable}[7]{r}{0.41\linewidth}
\vspace{-\intextsep}
\centering
\caption{Population prompt weights for Eq.~\eqref{eq:pop_pg}.}
\label{tab:online_algs}
\small
\setlength{\tabcolsep}{4pt}
\begin{tabular}{@{}lc@{}}
\toprule
Algorithm & Prompt weight $w_t(x)$\\
\midrule
REINFORCE, RLOO & $1$\\
GRPO & $1/\sqrt{p_t(x)(1-p_t(x))}$\\
MaxRL & $1/p_t(x)$\\
\bottomrule
\end{tabular}
\end{wraptable}

%% file: content/3_Recovery.tex
\section{Recovering Online RL from Its Rollouts}\label{sec:rollout_recovery}

\noindent\begin{minipage}{\linewidth}
\begin{wrapfigure}{r}{0.56\linewidth}
\vspace{-\intextsep}
\centering
\includegraphics[width=\linewidth]{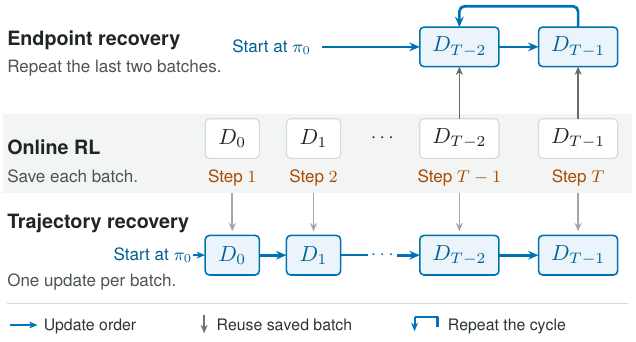}
\captionsetup{justification=raggedright}
\vspace{-5mm}
\caption{\textbf{Two replay protocols for a $T$-step rollout archive.} Both students start from $\pi_0$ and fit the saved responses by maximum likelihood without rewards. In this example, endpoint replay cycles over the final two batches to recover the performance of $\pi_T$. Trajectory replay takes one update per batch in chronological order.}
\label{fig:method}
\end{wrapfigure}
Given the rollouts generated during an online RL run, we investigate to what extent we can recover the performance of the original model.
Figure~\ref{fig:method} shows how the same archive can serve two goals. In \emph{trajectory recovery}, we seek to reproduce the model's performance throughout training. In \emph{endpoint recovery}, we seek only to recover its final performance.
We begin by investigating whether replaying the original RL updates can recover the training trajectory.
We then develop maximum likelihood replay, study the role of the order in which the rollouts are presented, and finally investigate how much of the training history is actually needed to recover the final model's performance.
\end{minipage}

\subsection{The Instability of Offline RL Replay}\label{sec:replay_instability}

A natural first attempt to recover an online RL run is to replay the original RL updates on the archived rollouts.
Starting from the same base model, $\hat\pi_0=\pi_0$, we process the batches in their original order, using the same update rule, rewards, and hyperparameters as the RL run which generated them.
With deterministic updates, replay would reproduce the online trajectory exactly. How closely does replay follow the original learning curve when numerical discrepancies and stochastic variation perturb the updates?

\paragraph{A population-level explanation.}
For a fixed prompt $x$, offline replay uses responses sampled from $\pi_t$ and evaluates their scores at the replayed policy $\hat\pi_t$. The replay counterpart of Eq.~\eqref{eq:pop_pg} is
\begin{equation}\label{eq:recovery_replay_gradient}
g_t^{\mathrm{replay}}(x)=w_t(x)\,\E_{y\sim\pi_t(\cdot\mid x)}
  \big[(r(x,y)-p_t(x))\,\nabla\log\hat\pi_t(y\mid x)\big].
\end{equation}
When $\hat\pi_t=\pi_t$, this is the online update. Small numerical discrepancies during training are one possible source of drift \citep{qi2025fp16}.
Once the policies differ, the score of $\hat\pi_t$ can have nonzero mean under $\pi_t$, so the baseline can contribute to the expected gradient \citep{degris2012offpolicy}.

For a prompt with $p_t(x)>0$, let $\pi_t^+(y\mid x)=\pi_t(y\mid x)r(x,y)/p_t(x)$ denote the distribution of responses conditioned on success.
If we hold the rollout-generating policy fixed at step $t$, Eq.~\eqref{eq:recovery_replay_gradient} gives the gradient for ascent on
\begin{equation}\label{eq:recovery_replay_kl}
w_t(x)p_t(x)\Bigl[
  \underbrace{-\mathrm{KL}(\pi_t^+\,\|\,\hat\pi_t)}_{
    \substack{\text{pull toward}\\\text{successful responses}}}
  \quad
  \underbrace{+\mathrm{KL}(\pi_t\,\|\,\hat\pi_t)}_{
    \substack{\text{baseline push away}\\\text{from the rollout policy}}}
\Bigr],
\end{equation}
where each policy is conditioned on $x$. Table~\ref{tab:online_algs} lists the prompt weight $w_t(x)$ used by each algorithm.
Under gradient ascent, the first term encourages the replayed policy to match the distribution of successful responses. The second term comes from the baseline and encourages divergence from the policy that generated the rollouts.
Its gradient vanishes on-policy. Once the policies differ, this term can amplify small discrepancies over successive updates.

\begin{figure}[H]
\centering
\includegraphics[width=\linewidth]{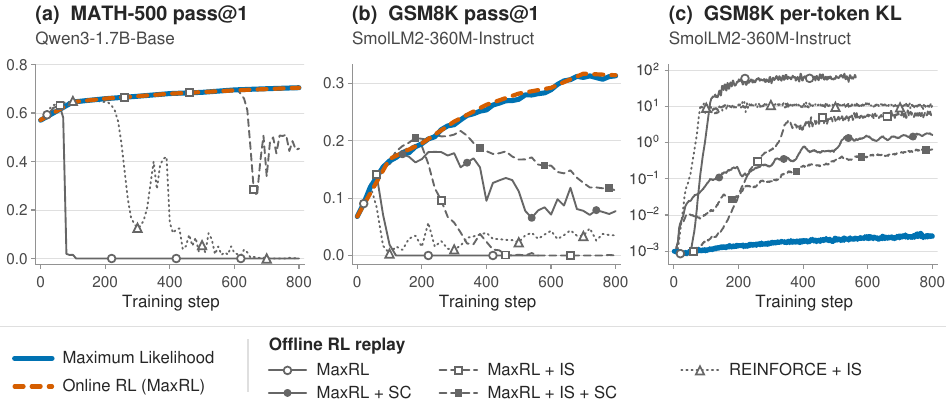}
\caption{\textbf{Maximum likelihood replay tracks online MaxRL as offline RL replay drifts.}
Panels (a) and (b) show pass@1 on MATH-500 and GSM8K. Panel (c) shows the per-token $\mathrm{KL}(\pi_t\,\|\,\hat\pi_t)$ from the rollout-generating policy to each replayed policy on GSM8K, on a logarithmic scale.
On MATH-500, we evaluate maximum likelihood replay and online MaxRL every 100 steps and offline RL replay every 10 steps. IS denotes truncated token-level importance sampling (Eq.~\eqref{eq:is_objective}). Filled markers indicate score centering (SC).}
\label{fig:replay_performance_and_drift}
\end{figure}

\paragraph{Empirical evidence.}
We replay the rollouts of an online MaxRL run of Qwen3-1.7B-Base~\citep{yang2025qwen3} on Polaris-53K~\citep{an2025polaris} and evaluate the replayed policy on MATH-500~\citep{hendrycks2021math,lightman2024verify} (Figure~\ref{fig:replay_performance_and_drift}(a)).
With uncorrected MaxRL replay, pass@1 initially improves, then falls to nearly zero by step~100.
With truncated token-level importance sampling, replay tracks online RL until roughly step~600, then drops sharply and only partially recovers.

We repeat the experiment with SmolLM2-360M-Instruct~\citep{allal2025smollm2} on GSM8K~\citep{cobbe2021gsm8k} and track pass@1 alongside the per-token divergence $\mathrm{KL}(\pi_t\,\|\,\hat\pi_t)$ (Figure~\ref{fig:replay_performance_and_drift}(b,c)).
With uncorrected MaxRL replay, the divergence grows by several orders of magnitude as performance collapses.
Score centering \citep{marek2026score} reduces drift, but even with importance sampling the replayed policy eventually falls behind online RL.

\subsection{Maximum Likelihood Replay Tracks the Online RL Learning Curve}\label{sec:recovery_learning_curve}
Can we recover the online trajectory by fitting each rollout-generating policy directly to its samples?
In \emph{maximum likelihood replay}, we use supervised fine-tuning (SFT) to maximize the log-likelihood over all archived responses, whether correct or incorrect.

\noindent\begin{minipage}{\linewidth}
\setlength{\parskip}{0.4em}
\begin{wrapfigure}{r}{0.40\linewidth}
\vspace{-\intextsep}
\centering
{\small\sffamily\bfseries Maximum Likelihood Replay Mechanism\par}
\includegraphics[width=\linewidth]{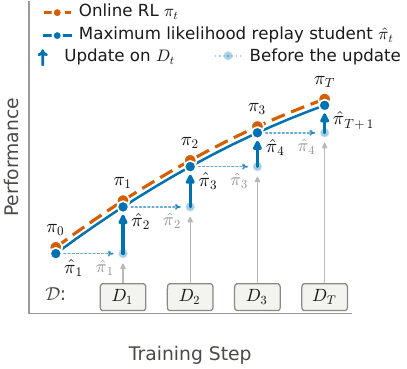}
\captionsetup{justification=raggedright}
\caption[Maximum Likelihood Replay Mechanism]{The online policy $\pi_t$ (dashed vermilion) generates batch $D_t$. The student (solid blue) takes one update on $D_t$ to fit $\pi_t$, yielding $\hat\pi_{t+1}$.}
\label{fig:sft_replay_mechanism}
\end{wrapfigure}
\paragraph{A population-level perspective.}
At step $t$, the online policy $\pi_t$ generates responses to prompts drawn from $\rho$. The population loss for a candidate policy $q$ is
\begin{equation}\label{eq:recovery_likelihood}
\begin{aligned}
\mathcal L_t(q)
&= -\E_{x\sim\rho,\,y\sim\pi_t(\cdot\mid x)}\big[\log q(y\mid x)\big] \\
&= \E_{x\sim\rho}\big[\mathrm{KL}\big(\pi_t(\cdot\mid x)\,\|\,q(\cdot\mid x)\big)\big] + \mathrm{Const.}
\end{aligned}
\end{equation}
For each prompt, maximum likelihood minimizes the same KL divergence that the baseline term in Eq.~\eqref{eq:recovery_replay_kl} encourages to increase.

If the student can represent $\pi_t$, an exact population fit gives $\hat\pi_{t+1}=\pi_t$ on prompts drawn from $\rho$, placing the student one step behind the online run (Figure~\ref{fig:sft_replay_mechanism}).
At initialization, for example, the population likelihood gradient is zero because the student already matches $\pi_0$. The first RL improvement becomes visible to the student in the next batch, sampled from $\pi_1$.

\emph{Can a single likelihood update on each finite batch keep pace as the online policy improves?}
\end{minipage}

\paragraph{Empirical evidence.}
Starting from the same base model, we take one maximum likelihood update on each archived batch in chronological order.
On MATH-500, the student's pass@1 tracks online MaxRL throughout training (Figure~\ref{fig:replay_performance_and_drift}(a)). On GSM8K, the student's per-token KL divergence from the rollout-generating policy stays small throughout training (Figure~\ref{fig:replay_performance_and_drift}(c)). Across four benchmarks, the student closely tracks the online learning curves at both pass@1 and pass@16 (Figure~\ref{fig:replay}).

\noindent\begin{minipage}{\linewidth}
Figure~\ref{fig:replay} compares the student's pass@1 and pass@16 with those of online MaxRL over 1,000 steps for Qwen3-1.7B-Base on MATH-500, AIME25~\citep{maa_aime}, GPQA-Diamond~\citep{rein2024gpqa}, and Knights and Knaves~\citep{xie2024memorization}.
On Knights and Knaves, the student follows online MaxRL from near-zero accuracy through the steep rise in performance. Across the four benchmarks, the mean absolute gap in pass@$k$, averaged over evaluation checkpoints and values of $k$, ranges from 0.2 to 1.1 percentage points (Table~\ref{tab:recovery_errors} in Appendix~\ref{sec:appendixreplay}).
The student tracks these gains using only the archived responses, with one update per batch and no reward labels.

\begin{figure}[H]
\centering
\includegraphics[width=\textwidth]{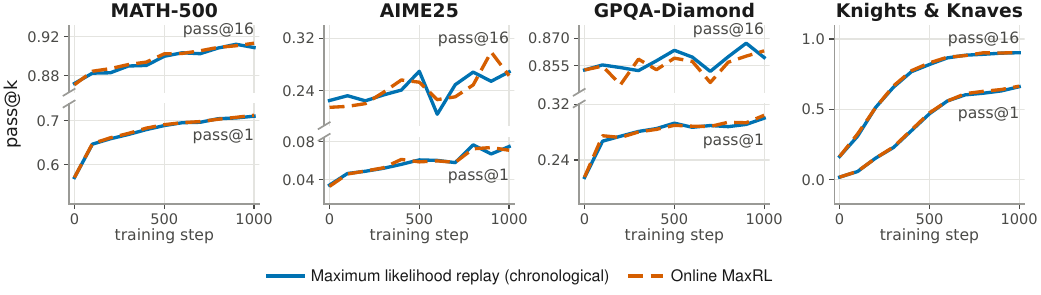}
\caption{\textbf{Chronological maximum likelihood replay follows the learning curve of online MaxRL.} Pass@1 and pass@16 over training of Qwen3-1.7B-Base, for replay in the original batch order (solid blue) and for online MaxRL (dashed vermilion). The $y$-axes are broken where the two metrics are far apart.}
\label{fig:replay}
\end{figure}
\end{minipage}

\noindent\begin{minipage}{\linewidth}
At the final checkpoint, chronological replay also closely matches online MaxRL across the evaluated values of pass@$k$ (Figure~\ref{fig:replay_end}).
This holds for Qwen3-1.7B-Base on all four benchmarks and for Qwen3-4B-Base~\citep{yang2025qwen3} on MATH-500, AIME25, and GPQA-Diamond.
The result generalizes: for both GRPO and RLOO, chronological maximum likelihood replay of the full rollout archive recovers the corresponding online run's final pass@$k$ on AIME 2024--2026 and MATH-500 with Qwen3-1.7B-Base (Figure~\ref{fig:different_online_algs} in Appendix~\ref{sec:app_online_algs}).

\begin{figure}[H]
\centering
\includegraphics[width=\textwidth]{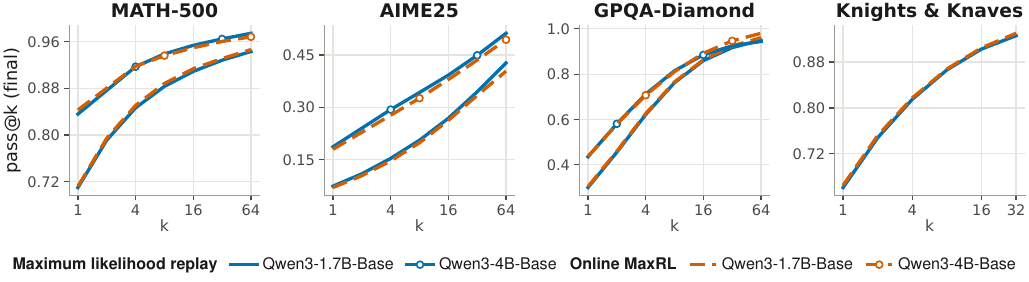}
\caption{\textbf{Chronological maximum likelihood replay recovers the final checkpoint of online MaxRL.} Final-checkpoint pass@$k$ for maximum likelihood replay in chronological order (solid blue) and online MaxRL (dashed vermilion). Open circles identify Qwen3-4B-Base. Unmarked curves show Qwen3-1.7B-Base. }
\label{fig:replay_end}
\end{figure}
\end{minipage}

\begin{takeaway}[Trajectory Recovery]
\textbf{Takeaway \thetcbcounter.} The learning trajectory of online RL can be recovered from its own rollouts through maximum likelihood, without access to rewards or intermediate model checkpoints.
\end{takeaway}

\subsection{The Role of Rollout Ordering}\label{sec:recovery_order}

Does maximum likelihood replay recover the online learning curve if the batch order changes? We present the same archive in chronological, reverse, and randomly shuffled order.

\paragraph{A population-level perspective.}
Let $\sigma$ be a permutation of the $T$ archived batches, so the student sees $D_{\sigma(s)}$ at replay step $s$. Under the exact population fit of Section~\ref{sec:recovery_learning_curve}, the student recovers the policy that generated that batch,
\begin{equation}\label{eq:recovery_order}
\hat{\pi}_{s+1}=\pi_{\sigma(s)},\qquad s=0,\ldots,T-1.
\end{equation}
Chronological replay therefore ends at $\pi_{T-1}$, the policy that generated the last online batch. Reverse replay ends at $\pi_0$, the policy that generated the first. Under a uniformly random ordering, the last batch is equally likely to come from any step of the online run.
How much does order matter when the student takes just one gradient update on each finite batch?

\paragraph{Empirical evidence.}
We replay the rollouts of an online MaxRL run of Qwen3-1.7B-Base on Polaris-53K. All three students start from the same base model, see every rollout once, and differ only in the order of the batches. We compare pass@1 throughout training on MATH-500, AIME25, Minerva~\citep{lewkowycz2022minerva}, and GPQA-Diamond (Figure~\ref{fig:order}) and the error at the final checkpoint (Figure~\ref{fig:order_error}).

\noindent\begin{minipage}{\linewidth}
On MATH-500, reverse replay raises pass@1 from about 57\% to 70\% in the first 100 updates, close to the final performance of online MaxRL. As the student encounters progressively earlier batches, pass@1 falls back to about 57\%. Chronological replay tracks online MaxRL, while shuffled replay levels off below it (Figure~\ref{fig:order}).

\begin{figure}[H]
\centering
\includegraphics[width=\textwidth]{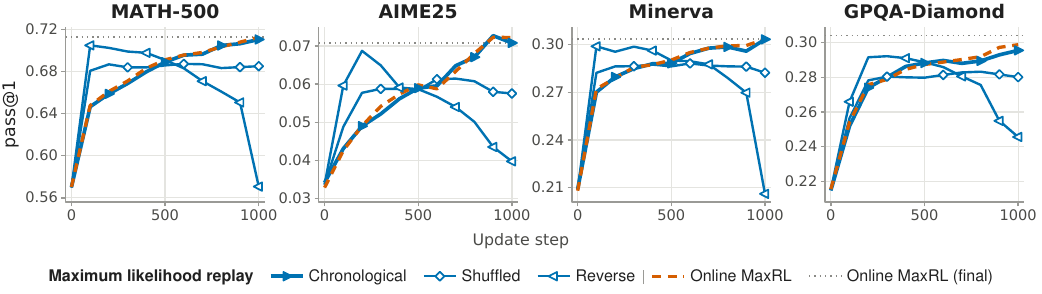}
\caption{\textbf{The same rollouts produce different learning curves when replayed in different orders.} Each maximum likelihood replay run starts from Qwen3-1.7B-Base and sees every batch from the Polaris-53K MaxRL run once. On the blue replay curves, triangles point right for chronological order and left for reverse order. Diamonds mark shuffled replay. The dashed vermilion curve shows online MaxRL, and the dotted gray line marks its final pass@1. On AIME25 and GPQA-Diamond, step zero shows each run's unsmoothed score. Later points use a centered moving average over up to three evaluations.}
\label{fig:order}
\end{figure}
\end{minipage}

\noindent\begin{minipage}{\linewidth}
On each benchmark, the \emph{endpoint error} $\epsilon_{\mathrm{end}}$ is the absolute gap in pass@$k$ between the model under evaluation and the final online MaxRL checkpoint, averaged over $k\in\{1,2,4,8,16,32\}$ (Eq.~\eqref{eq:end}). Across all four benchmarks, chronological replay has the smallest error, shuffled replay is intermediate, and reverse replay has the largest (Figure~\ref{fig:order_error}). On MATH-500 and AIME25, reverse replay ends slightly farther from online MaxRL than the base model does.

\begin{figure}[H]
\centering
\includegraphics[width=\textwidth]{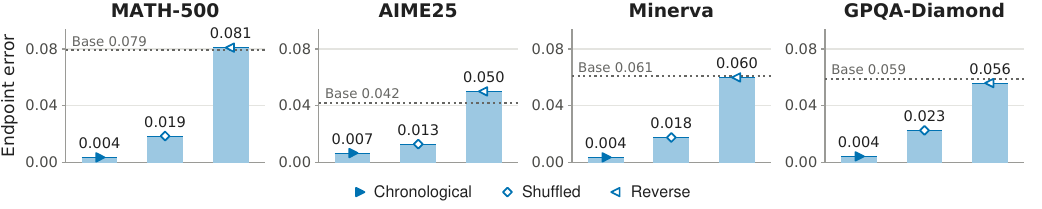}
\caption{\textbf{Chronological replay has the lowest endpoint error on all four benchmarks.} Bar height gives the endpoint error after 1,000 replay updates. As in Figure~\ref{fig:order}, right-pointing triangles mark chronological replay, diamonds mark shuffled replay, and left-pointing triangles mark reverse replay. Each dotted gray line marks the base model's endpoint error on that benchmark before replay.}
\label{fig:order_error}
\end{figure}
\end{minipage}

Even after the student has seen every rollout, its final performance depends on the order in which it saw them. In reverse replay, the student approaches the final MATH-500 score while it is still training on late-stage rollouts. 

\subsection{Recovering Final Performance from Few Rollouts}\label{sec:recovery_few_rollouts}

\emph{If final performance is the goal, how much of the earlier training history do we need?}
The previous results suggest that the performance of the replayed
model is largely determined by the distribution of the rollouts it
is currently learning from. If we are only interested in recovering
the final performance of online RL, perhaps we do not need to
replay its entire training history.

\paragraph{A population-level perspective.}
Recall that maximum likelihood on samples generated by $\pi_t$
recovers $\pi_t$ at the population optimum. Thus, in principle,
access to the last rollout-generating distribution $\pi_{T-1}$
makes the entire preceding training history unnecessary.

In practice, however, we only have access to a finite number of
samples from this distribution. To increase the amount of available
data, we can use the rollouts generated during the last $m$ RL
steps. Assigning equal weight to these steps, the corresponding
population objective is
\begin{equation}
    \min_q \;
    \frac{1}{m}
    \sum_{t=T-m}^{T-1}
    \mathbb{E}_{x\sim\rho}
    \left[
        \mathrm{KL}\big(
        \pi_t(\cdot|x)\,\|\,q(\cdot|x)
        \big)
    \right].
\end{equation}
The unrestricted minimizer of this objective is simply the mixture
\begin{equation}
    q^\star(\cdot|x)
    =
    \frac{1}{m}
    \sum_{t=T-m}^{T-1}
    \pi_t(\cdot|x).
\end{equation}

Provided that the policy changes sufficiently little toward the end
of training, this mixture is close to the final RL policy.
This raises an empirical question: \emph{how many of the rollouts
generated during RL are actually needed to recover its final
performance?}

\noindent\begin{minipage}{\linewidth}
\paragraph{Empirical evidence.}
Starting from the same base model, we perform repeated maximum
likelihood updates on the rollouts generated during the last $m$
RL steps. Remarkably, for Qwen3-1.7B-Base, the rollouts of a single
RL step are sufficient to closely recover the performance acquired
over a thousand RL steps, across mathematical reasoning and
Knights and Knaves (Figure~\ref{fig:onebatch}). Thus, recovering the final performance
requires only $0.1\%$ of the distinct rollouts generated during
online RL in these settings.
On Knights and Knaves, the final pass@$k$ gap from online MaxRL, averaged over $k$, is 71.9 percentage points for the base model, 0.9 points after replay on the final batch alone, and 0.3 points after replay on the full archive (Table~\ref{tab:last_step_errors} in Appendix~\ref{sec:appendixreplay}).

A single batch is not sufficient in every setting.
For SmolLM2-360M-Instruct on GSM8K, increasing the number of
archived batches progressively closes the gap with online RL,
with the last $50$ batches recovering most of its performance.

\begin{figure}[H]
\centering
\includegraphics[width=\textwidth]{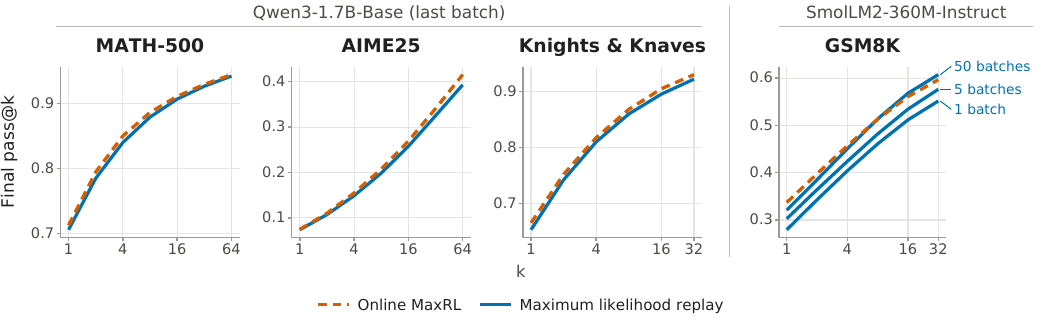}
\caption{\textbf{Replaying the final batch nearly recovers online MaxRL performance on all three Qwen benchmarks.}
Repeated maximum likelihood updates from the base model on the last of 1,000 batches yield an endpoint error below one percentage point on each benchmark.
On GSM8K, SmolLM2's endpoint error falls from 5.2 to 3.0 to 0.9 percentage points using the last 1, 5, and 50 batches, respectively.}
\label{fig:onebatch}
\end{figure}
\end{minipage}

\noindent\begin{minipage}{\linewidth}
We further investigate how many rollouts from the last RL step
are actually needed (Figure~\ref{fig:budget}). On some benchmarks, even a small
subset of these rollouts suffices to closely recover the final
performance.

\par\smallskip
\captionsetup{type=figure}
\includegraphics[width=\textwidth]{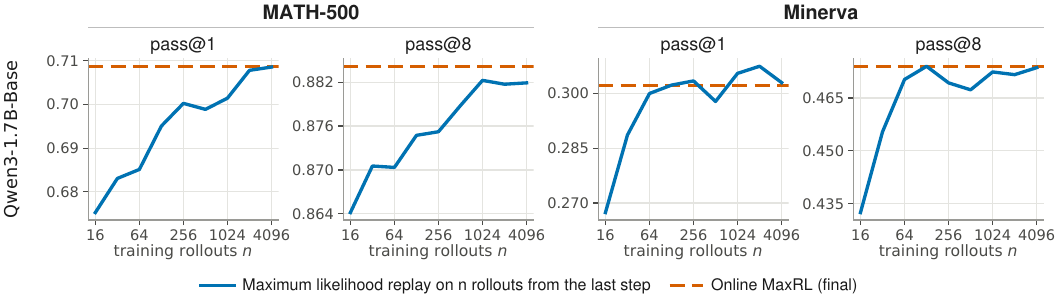}
\caption{\textbf{Minerva approaches full-batch replay with fewer rollouts than MATH-500.} Each pair shows pass@1 and pass@8 after 100 maximum likelihood replay updates from Qwen3-1.7B-Base on $n$ randomly selected rollouts from the final Polaris-53K batch of 4,096 (solid blue). Dashed vermilion lines show final online MaxRL performance.}
\label{fig:budget}
\end{minipage}

These findings suggest that much of the computational effort in RL
goes into discovering an improved sampling distribution rather than
learning it from samples. Once samples from this distribution are
available, the capabilities acquired through RL can often be
recovered through maximum likelihood replay on a remarkably small fraction
of the original training data.

\begin{takeaway}[Endpoint Recovery]
\textbf{Takeaway \thetcbcounter.} Discovering an improved sampling distribution through RL can require substantially more data than learning from it. The rollouts of a few late-stage RL steps suffice to recover the performance acquired over a thousand steps.
\end{takeaway}

%% file: content/4_Applications.tex
\section{Further Applications of Archived Rollouts}\label{sec:rollout_applications}

The previous section establishes that much of what a model learns
during RL can be recovered from the rollouts it generates. This
opens up possibilities for reusing the compute invested in RL
beyond recovering the original model. Here, we investigate whether
the capabilities of independently trained models can be combined
and whether the capabilities acquired by one model can be
transferred to another, using only their archived rollouts.

\subsection{Combining RL Specialists in Data Space}\label{sec:rollout_specialists}

Consider several models independently trained through RL to acquire
different capabilities. A common approach to combining these
capabilities is to merge the weights of the resulting models.
However, since the capabilities acquired during RL are also
reflected in the rollouts, an alternative is to combine the
models in data space.

\paragraph{A population-level perspective.}
Consider $M$ specialists trained from the same base model on
different prompt distributions $\rho_1,\ldots,\rho_M$.
Let $\pi_t^k$ denote the policy of specialist $k$ at training
step $t$. At step $t$, maximum likelihood on the pooled rollouts
gives the population optimization problem
\begin{equation}
    \min_q
    \sum_{k=1}^{M}
    w_k\,
    \mathbb{E}_{x\sim\rho_k}
    \left[
        \mathrm{KL}\big(
        \pi_t^k(\cdot|x)\,\|\,q(\cdot|x)
        \big)
    \right],
\end{equation}
where $w_k>0$ determines the relative contribution of each
specialist. For each prompt $x$ in the pooled support, the
unrestricted optimum is a mixture of the specialists' response
distributions, with weights proportional to $w_k\rho_k(x)$.

If the prompt distributions have disjoint supports and the model
class is sufficiently expressive, a minimizer of this objective
simultaneously matches each specialist on its respective domain.
Thus, in principle, a single model could track the evolution of
several independently trained RL policies.

Furthermore, our findings in Section~\ref{sec:recovery_few_rollouts} suggest that combining
the final capabilities of the specialists may require only a
small fraction of their rollouts.

\paragraph{Empirical evidence.}
We train one student on the archives of two MaxRL specialists trained from
Qwen3-1.7B-Base, one on Polaris-53K and one on Knights and Knaves.
Starting from the same base model, the student takes one maximum likelihood
update on the math batch at each online RL step, then one on the puzzle batch
from that step. Figure~\ref{fig:compose_curves} shows that the student
closely tracks both specialists on their respective benchmarks.

\par\smallskip
\noindent\begin{minipage}{\linewidth}
\captionsetup{type=figure}
\centering
\includegraphics[width=\linewidth]{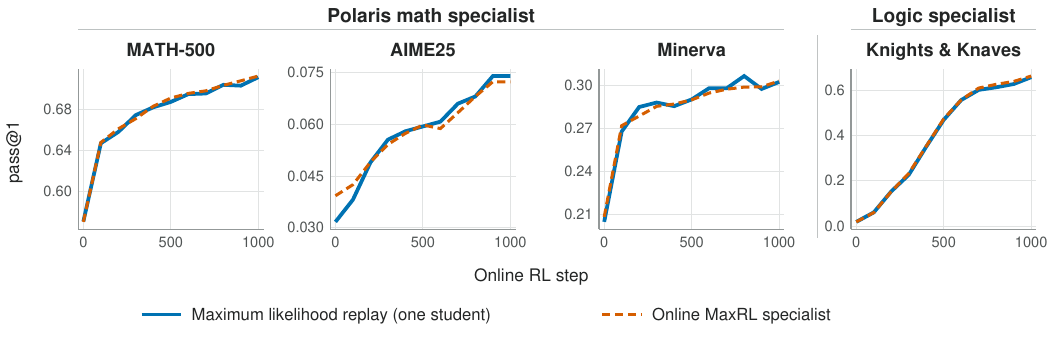}
\caption{\textbf{One replay student tracks two online MaxRL specialists.}
The solid blue curve shows the same student's pass@1 in all four panels.
Dashed vermilion shows the math specialist on the first three benchmarks
and the logic specialist on Knights and Knaves. Each online RL step corresponds
to two student updates. AIME25 is smoothed over three evaluations.}
\label{fig:compose_curves}
\end{minipage}

\begin{figure}[!htb]
\centering
\includegraphics[width=\linewidth]{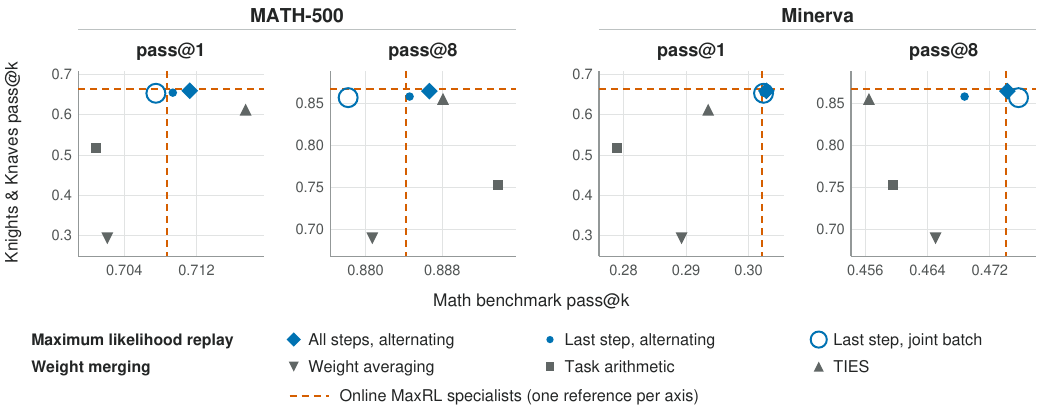}
\caption{\textbf{Replay on the final batch from each specialist retains performance on both tasks.}
Each point is one model evaluated at the same $k$ on the indicated math benchmark
(horizontal axis) and on Knights and Knaves (vertical axis). The dashed lines
intersect at the scores of the two online MaxRL specialists on their respective tasks.
Blue diamonds show maximum likelihood replay on both full archives. Blue circles
use only the two final batches, alternating between tasks (filled) or combining
them in each update (open). Gray markers show weight merges of the final checkpoints.}
\label{fig:composition}
\end{figure}

We also investigate whether their final capabilities can be
combined using only the rollouts of their last RL steps.
Figure~\ref{fig:composition} shows that this procedure retains most of the
performance of both specialists and generally provides a
better trade-off between their capabilities than weight-space
merging. Table~\ref{table:composition_math_knk} in Appendix~\ref{sec:compose_protocol}
reports additional replay windows and results on
AIME 2024--2026~\citep{maa_aime} and BeyondAIME~\citep{seed2025thinking}.

These results suggest that archived rollouts provide an
alternative representation through which the capabilities
acquired during independent RL runs can be combined.

\FloatBarrier
\subsection{Transferring Capabilities Across Models}\label{sec:rollout_transfer}

So far, we have recovered the capabilities acquired during RL
by starting from the same base model. However, since maximum
likelihood operates on the data generated during RL rather than
on the parameters of the model, the same rollouts could also
be used to train a different model.

\paragraph{A population-level perspective.}
Let $\mathcal{Q}$ denote the class of policies that the new model
can represent. Maximum likelihood on the rollouts of the last
$m$ RL steps amounts, at the population level, to solving
\begin{equation}
    \min_{q\in\mathcal{Q}}
    \frac{1}{m}
    \sum_{t=T-m}^{T-1}
    \mathbb{E}_{x\sim\rho}
    \left[
        \mathrm{KL}\big(
        \pi_t(\cdot|x)\,\|\,q(\cdot|x)
        \big)
    \right].
\end{equation}
By Section~\ref{sec:recovery_few_rollouts}, replay finds the best
approximation in $\mathcal{Q}$, under expected forward KL, to the
average of the last $m$ rollout-generating policies. Sequence-level
distillation from the final RL checkpoint finds the best approximation
in the same class to $\pi_T$ by solving
\begin{equation}
    \min_{q\in\mathcal{Q}}
    \mathbb{E}_{x\sim\rho}
    \left[
        \mathrm{KL}\big(
        \pi_T(\cdot|x)\,\|\,q(\cdot|x)
        \big)
    \right].
\end{equation}
For $m=1$, replay targets $\pi_{T-1}$, the policy that generated
the last batch, and distillation targets $\pi_T$, the policy after
the final RL update. Replay uses data already generated during RL,
without requiring access to the final model or generating additional
rollouts.

\paragraph{Empirical evidence.}
We investigate this possibility by training Qwen3-0.6B-Base~\citep{yang2025qwen3}
on the archived rollouts of an online MaxRL run of
Qwen3-1.7B-Base. We compare chronological maximum likelihood replay
on the entire rollout archive, repeated maximum likelihood
updates on the rollouts of the last RL step alone, and
conventional distillation from the final RL checkpoint.

\noindent\begin{minipage}{\linewidth}
Figure~\ref{fig:transfer} shows that both forms of replay closely
track the distillation pass@$k$ curves on MATH-500 and Minerva,
even when the student trains on a single archived batch without
distinguishing between correct and incorrect responses. At larger
$k$, last-step replay falls below distillation on AIME25 and
BeyondAIME, while full-archive replay exceeds distillation on
BeyondAIME.

\begin{figure}[H]
\centering
\includegraphics[width=\textwidth]{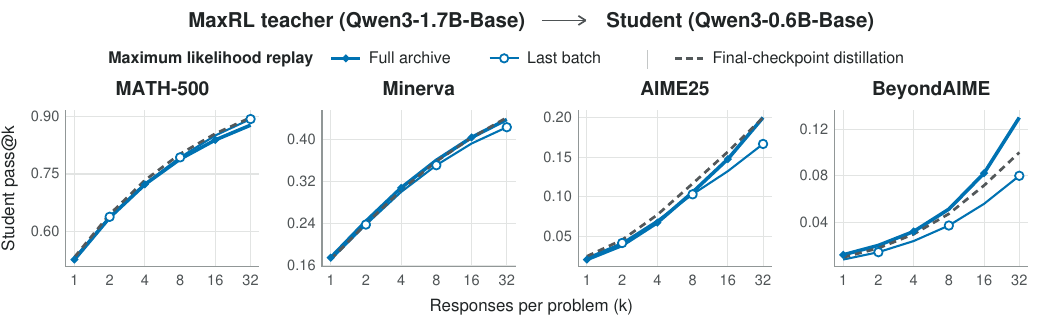}
\caption{\textbf{Cross-model transfer by maximum likelihood replay.}
All curves show pass@$k$ for a Qwen3-0.6B-Base student. Blue diamonds show
replay on the full rollout archive of a Qwen3-1.7B-Base teacher trained with
MaxRL on Polaris-53K. Open blue circles show replay on its last batch alone.
The dashed gray curve shows distillation on fresh samples from the final
teacher checkpoint.}
\label{fig:transfer}
\end{figure}
\end{minipage}

A smaller student can therefore learn from responses already
generated during the larger model's RL run, even when its final
checkpoint is unavailable.

\begin{takeaway}[Reusability]
\textbf{Takeaway \thetcbcounter.} RL rollouts are reusable assets: capabilities acquired through independent RL runs can be combined in data space and transferred across models.
\end{takeaway}

%% file: content/5_Related_Work.tex
\section{Related Work}

\paragraph{Offline RL and rollout replay.}
Experience replay is a central component of value-based reinforcement learning~\citep{mnih2015human}, whereas policy-gradient methods generally require corrections when learning from off-policy data~\citep{degris2012offpolicy,levine2020offline}.
Recent work has investigated the instability arising from discrepancies between training and rollout-generating policies, particularly in asynchronous RL and under training-inference mismatch~\citep{noukhovitch2025asynchronous,marek2026score,bai2026regularized}.
We study a completed online RL run whose original rollout batches are available in training order. We ask whether those batches suffice to recover the run's learning curve through maximum likelihood replay.

\paragraph{Policy distillation and the role of RL.}
Distilling RL-trained policies through supervised learning dates back to \citet{rusu2016policy}, who also investigate tracking an evolving policy during Q-learning using its action-value predictions.
More recently, \citet{chen2025towards} show that a small set of correct responses, including responses drawn from an RL replay buffer, can recover much of an RL-trained model's performance.
\citet{lu2025onpolicy} argue that much of the computation in RL goes into discovering effective strategies, which can subsequently be learned more efficiently through on-policy distillation.
Our findings complement these observations. A single maximum likelihood update per archived batch, in chronological order, closely tracks the original RL learning curve without access to rewards or teacher checkpoints.

\paragraph{Reusing RL training data.}
Self-training methods typically fine-tune models on successful responses sampled from their own distributions~\citep{zelikman2022star,singh2024beyond,dong2023raft}.
Recent work has investigated reusing RL rollouts for expert consolidation through offline RL~\citep{chen2026regen} and for further improving an already trained model through selective supervision~\citep{zhang2026ross}.
We investigate how much of the knowledge acquired during RL can be recovered from its unfiltered rollouts through maximum likelihood replay, starting from the original base model.

\paragraph{Combining model capabilities.}
Weight-space methods combine independently trained models by averaging their parameters~\citep{wortsman2022modelsoups}, combining task vectors~\citep{ilharco2023taskarithmetic}, or resolving parameter interference~\citep{yadav2023ties}.
Alternatively, multi-teacher distillation combines the capabilities of independently trained specialists through supervision from their output distributions~\citep{rusu2016policy,ma2026mopd,gao2026openmopd}.
Using only archived rollouts, a single student can track both specialists' learning curves. In our experiments, training on the last rollout batch from each specialist already recovers most of its final performance on its own task.

%% file: content/6_Conclusion.tex
\section{Conclusion}
Our findings suggest a distinction between two roles of computation in reinforcement learning: discovering an improved sampling distribution and learning from it. While these processes are intertwined in online RL, they need not be repeated together. Once a useful distribution has been discovered, the data it produces can serve as a vehicle for transferring the benefits of that discovery, potentially at a much lower computational cost. From this perspective, the computational investment in RL need not be tied to the lifetime or architecture of a particular model. Rather than viewing RL solely as a process for producing better models, we may also view it as a process for discovering better training distributions, whose value can extend well beyond the original training run.

%% file: tables/recovery_errors_combined.tex
\begin{table}[H]
\centering
\caption{Recovery errors of maximum likelihood replay against online MaxRL. (a) Chronological replay on all rollouts, on the benchmarks of Figure~\ref{fig:replay}. $\epsilon_{\mathrm{traj}}$ (Eq.~\eqref{eq:traj}) is over steps $0,100,\ldots,1000$, and $\epsilon_{\mathrm{end}}$ (Eq.~\eqref{eq:end}) is at step 1000. (b) Endpoint error $\epsilon_{\mathrm{end}}$ of replay on all rollouts and on the rollouts of the last $m$ steps. In both tables, the base model uses the untrained base model $\pi_0$ in place of the student at every step. For Qwen3-4B-Base, $\mathcal K$ omits $k=2$.}
\label{tab:recovery_combined}
\small
\setlength{\tabcolsep}{2.5pt}
\begin{subtable}[c]{0.485\textwidth}
\centering
\caption{Trajectory and endpoint recovery}
\label{tab:recovery_errors}
\begin{tabular}{lccc}
\toprule
 & \multicolumn{2}{c}{Qwen3-1.7B-Base} & Qwen3-4B-Base\\
\cmidrule(lr){2-3}\cmidrule(lr){4-4}
Benchmark & $\epsilon_{\mathrm{traj}}$ & $\epsilon_{\mathrm{end}}$ & $\epsilon_{\mathrm{end}}$\\
\midrule
\multicolumn{4}{@{}l}{\emph{Replay on all rollouts}}\\
MATH-500 & 0.002 & 0.004 & 0.004\\
AIME25 & 0.011 & 0.007 & 0.013\\
GPQA-Diamond & 0.004 & 0.004 & 0.008\\
KnK & 0.006 & 0.003 & --\\
\midrule
\multicolumn{4}{@{}l}{\emph{Base model}}\\
MATH-500 & 0.055 & 0.079 & 0.068\\
AIME25 & 0.025 & 0.042 & 0.060\\
GPQA-Diamond & 0.043 & 0.059 & 0.050\\
KnK & 0.490 & 0.719 & --\\
\bottomrule
\end{tabular}
\end{subtable}\hfill
\begin{subtable}[c]{0.495\textwidth}
\centering
\caption{Endpoint recovery from the last $m$ steps}
\label{tab:last_step_errors}
\begin{tabular}{llccc}
\toprule
Model & Benchmark & \begin{tabular}{@{}c@{}}All\\rollouts\end{tabular} & \begin{tabular}{@{}c@{}}Last $m$\\steps\end{tabular} & \begin{tabular}{@{}c@{}}Base\\model\end{tabular}\\
\midrule
\multirow{3}{*}{\begin{tabular}{@{}l@{}}Qwen3-1.7B-Base\\($m=1$)\end{tabular}} & MATH-500 & 0.004 & 0.007 & 0.079\\
 & AIME25 & 0.007 & 0.008 & 0.042\\
 & KnK & 0.003 & 0.009 & 0.719\\
\midrule
\begin{tabular}{@{}l@{}}SmolLM2-360M\\($m=50$)\end{tabular} & GSM8K & 0.003 & 0.009 & 0.226\\
\bottomrule
\end{tabular}
\end{subtable}
\end{table}

%% file: tables/math_knk_results.tex
\begin{table}[H]
\centering
\caption{Composing Math and KnK tasks: pass@1/8/32 (\%) across six math benchmarks, their weighted-average pool (\emph{All Math}), and Knights-and-Knaves. Among the six merge and maximum likelihood replay checkpoints (i.e., excluding the base model and the two online-maxRL specialists), \textbf{bold} marks the best value in each column at each $k$, \underline{underline} the second best, and \textit{italic} the third best; tied values share the same mark. \colorbox{yellow!25}{Shaded} cells exceed the corresponding online-maxRL specialist (Polaris-53K for the math columns, KnK for the KnK column) at the same $k$.}
\label{tab:pareto-front}
\resizebox{\textwidth}{!}{%
\begin{tabular}{llcccccccc}
\toprule
Checkpoint & $k$ & MATH500 & Minerva & BeyondAIME & AIME24 & AIME25 & AIME26 & All Math & KnK \\
\midrule[1.5pt]
\multicolumn{10}{c}{Base model and Online-MaxRL final checkpoints}\\
\midrule
\multirow{3}{*}{Qwen3-1.7B-Base} & pass@1 & 57.4 & 21.1 & 1.2 & 4.0 & 3.1 & 1.7 & 36.2 & 2.1 \\
 & pass@8 & 82.3 & 43.1 & 6.4 & 14.9 & 15.4 & 7.1 & 56.8 & 12.3 \\
 & pass@32 & 89.2 & 52.2 & 14.0 & 26.4 & 29.3 & 10.0 & 64.6 & 27.4 \\
\midrule
\multirow{3}{*}{Online MaxRL (Polaris-53K)} & pass@1 & 70.8 & 30.5 & 1.6 & 11.6 & 7.4 & 6.4 & 46.4 & 3.4 \\
 & pass@8 & 88.5 & 47.9 & 8.7 & 25.9 & 20.9 & 18.6 & 62.5 & 16.2 \\
 & pass@32 & 93.3 & 56.2 & 17.0 & 33.1 & 33.8 & 27.6 & 69.1 & 31.3 \\
\midrule
\multirow{3}{*}{Online MaxRL (KnK)} & pass@1 & 67.4 & 27.4 & 1.5 & 8.8 & 5.4 & 3.8 & 43.5 & 69.7 \\
 & pass@8 & 86.1 & 45.1 & 9.0 & 25.3 & 18.3 & 14.0 & 60.2 & 88.9 \\
 & pass@32 & 92.6 & 54.8 & 20.0 & 33.3 & 33.3 & 23.3 & 68.5 & 94.5 \\
\midrule[1.5pt]
\multicolumn{10}{c}{Weight-space composition}\\
\midrule
\multirow{3}{*}{Averaged (50\% math)} & pass@1 & 70.1 & 29.0 & 1.4 & 8.4 & 6.3 & 4.6 & 45.4 & 30.2 \\
 & pass@8 & 87.7 & 45.4 & 8.1 & 22.7 & \textit{19.7} & 17.3 & 61.1 & 71.3 \\
 & pass@32 & 92.0 & 52.2 & \textit{17.0} & \underline{30.0} & \underline{33.3} & \textit{26.7} & 67.2 & 87.6 \\
\midrule
\multirow{3}{*}{Task-arithmetic} & pass@1 & 70.4 & 28.2 & 1.3 & \textit{11.2} & \underline{6.9} & \textbf{6.3} & 45.5 & 54.3 \\
 & pass@8 & \cellcolor{yellow!25}\textbf{88.9} & 45.4 & 7.4 & 25.4 & \cellcolor{yellow!25}\textbf{21.0} & \cellcolor{yellow!25}\textbf{23.4} & 62.0 & 78.6 \\
 & pass@32 & \underline{93.2} & 53.7 & \textit{17.0} & \underline{30.0} & \cellcolor{yellow!25}\textbf{36.7} & \cellcolor{yellow!25}\textbf{33.3} & \textit{68.5} & 86.9 \\
\midrule
\multirow{3}{*}{TIES-merged (top 20\%)} & pass@1 & \cellcolor{yellow!25}\underline{71.2} & 29.1 & \textit{1.5} & 10.7 & 6.2 & 5.5 & \textit{46.1} & 64.1 \\
 & pass@8 & \cellcolor{yellow!25}\underline{88.6} & \textit{46.4} & 7.9 & \cellcolor{yellow!25}\textbf{27.0} & 16.1 & \cellcolor{yellow!25}\textit{19.1} & 61.9 & \underline{88.4} \\
 & pass@32 & \textit{92.8} & \textit{55.1} & 16.0 & \underline{30.0} & 23.3 & \cellcolor{yellow!25}\underline{30.0} & 68.1 & \cellcolor{yellow!25}\textbf{95.5} \\
\midrule[1.5pt]
\multicolumn{10}{c}{Data-space composition (ours)}\\
\midrule
\multirow{3}{0.22\linewidth}{Last-step alternating replay (epoch 200)} & pass@1 & 70.8 & \cellcolor{yellow!25}\textbf{30.7} & \cellcolor{yellow!25}\textbf{1.7} & \cellcolor{yellow!25}\underline{11.7} & \underline{6.9} & \textbf{6.3} & \underline{46.4} & \underline{68.3} \\
 & pass@8 & 88.2 & \textbf{47.9} & \cellcolor{yellow!25}\underline{9.2} & \cellcolor{yellow!25}\textit{26.0} & 18.2 & \cellcolor{yellow!25}\underline{20.0} & \textbf{62.4} & \textit{88.1} \\
 & pass@32 & \textit{92.8} & \underline{56.2} & \textit{17.0} & \cellcolor{yellow!25}\textbf{33.3} & \textit{30.0} & \cellcolor{yellow!25}\textbf{33.3} & \underline{68.9} & \underline{94.1} \\
\midrule
\multirow{3}{0.22\linewidth}{Last-25\%-steps alternating replay} & pass@1 & \cellcolor{yellow!25}\textbf{71.4} & \textit{30.1} & \underline{1.6} & 10.6 & \textbf{7.2} & \underline{6.2} & \cellcolor{yellow!25}\textbf{46.5} & \textit{68.0} \\
 & pass@8 & \textit{88.5} & \underline{47.0} & \cellcolor{yellow!25}\textit{8.9} & 25.2 & \underline{20.5} & \cellcolor{yellow!25}18.8 & \textit{62.2} & 87.9 \\
 & pass@32 & \cellcolor{yellow!25}\textbf{93.6} & 54.8 & \cellcolor{yellow!25}\underline{18.0} & \cellcolor{yellow!25}\textbf{33.3} & \underline{33.3} & \textit{26.7} & \underline{68.9} & 93.3 \\
\midrule
\multirow{3}{0.22\linewidth}{Last-50\%-steps alternating replay} & pass@1 & \cellcolor{yellow!25}\textit{71.1} & \underline{30.3} & \cellcolor{yellow!25}\textbf{1.7} & \cellcolor{yellow!25}\textbf{12.5} & \textit{6.8} & \textit{6.0} & \cellcolor{yellow!25}\textbf{46.5} & \textbf{68.8} \\
 & pass@8 & 88.4 & \underline{47.0} & \cellcolor{yellow!25}\textbf{10.0} & \cellcolor{yellow!25}\underline{26.8} & 19.4 & 17.2 & \underline{62.3} & \textbf{88.7} \\
 & pass@32 & 92.6 & \cellcolor{yellow!25}\textbf{56.6} & \cellcolor{yellow!25}\textbf{23.0} & \cellcolor{yellow!25}\textbf{33.3} & \textit{30.0} & \cellcolor{yellow!25}\underline{30.0} & \cellcolor{yellow!25}\textbf{69.4} & \textit{93.9} \\
\bottomrule
\end{tabular}%
}
\label{table:composition_math_knk}
\end{table}

%% file: tables/math_knk_token_balanced.tex
\begin{table}[H]
\centering
\caption{Composing Math and KnK tasks with balanced tokens: pass@1/8/32 (\%) across six math benchmarks, their weighted-average pool (\emph{All Math}), and Knights-and-Knaves. Among the six merge and maximum likelihood replay checkpoints (i.e., excluding the base model and the two online-maxRL specialists), \textbf{bold} marks the best value in each column at each $k$, \underline{underline} the second best, and \textit{italic} the third best; tied values share the same mark. \colorbox{yellow!25}{Shaded} cells exceed the corresponding online-maxRL specialist (Polaris-53K for the math columns, KnK for the KnK column) at the same $k$.}
\label{tab:math_knk_token_balanced}
\resizebox{\textwidth}{!}{%
\begin{tabular}{llcccccccc}
\toprule
Checkpoint & $k$ & MATH500 & Minerva & BeyondAIME & AIME24 & AIME25 & AIME26 & All Math & KnK \\
\midrule[1.5pt]
\multicolumn{10}{c}{Last-step}\\
\midrule
\multirow{3}{0.22\linewidth}{Last-step alternating replay} & pass@1 & 70.8 & \cellcolor{yellow!25}\underline{30.7} & \cellcolor{yellow!25}\textit{1.7} & \cellcolor{yellow!25}11.7 & \textit{6.9} & \underline{6.3} & 46.4 & \textit{68.3} \\
 & pass@8 & 88.2 & \underline{47.9} & \cellcolor{yellow!25}9.2 & \cellcolor{yellow!25}\underline{26.0} & 18.2 & \cellcolor{yellow!25}\textbf{20.0} & \textit{62.4} & \textit{88.1} \\
 & pass@32 & \textit{92.8} & \textit{56.2} & 17.0 & \cellcolor{yellow!25}\textbf{33.3} & \textit{30.0} & \cellcolor{yellow!25}\textbf{33.3} & \textit{68.9} & \textbf{94.1} \\
\midrule
\multirow{3}{0.22\linewidth}{Last-step alternating replay (token-balanced)} & pass@1 & 70.4 & \textit{30.5} & \cellcolor{yellow!25}\underline{1.8} & \cellcolor{yellow!25}\underline{11.9} & 6.2 & 5.7 & 46.2 & 68.0 \\
 & pass@8 & 87.9 & \cellcolor{yellow!25}\textbf{48.0} & \cellcolor{yellow!25}\textit{9.3} & 24.2 & 18.1 & \cellcolor{yellow!25}\underline{19.4} & 62.1 & 87.8 \\
 & pass@32 & \textit{92.8} & 54.8 & \cellcolor{yellow!25}\textit{18.0} & \underline{30.0} & 23.3 & \cellcolor{yellow!25}\textbf{33.3} & 68.3 & \underline{93.9} \\
\midrule[1.5pt]
\multicolumn{10}{c}{Last-25\%-steps}\\
\midrule
\multirow{3}{0.22\linewidth}{Last-25\%-steps alternating replay} & pass@1 & \cellcolor{yellow!25}\underline{71.4} & 30.1 & 1.6 & 10.6 & \underline{7.2} & \textit{6.2} & \cellcolor{yellow!25}\textit{46.5} & 68.0 \\
 & pass@8 & \underline{88.5} & \textit{47.0} & \cellcolor{yellow!25}8.9 & 25.2 & \underline{20.5} & \cellcolor{yellow!25}\textit{18.8} & 62.2 & 87.9 \\
 & pass@32 & \cellcolor{yellow!25}\textbf{93.6} & 54.8 & \cellcolor{yellow!25}\textit{18.0} & \cellcolor{yellow!25}\textbf{33.3} & \underline{33.3} & \textit{26.7} & \textit{68.9} & 93.3 \\
\midrule
\multirow{3}{0.22\linewidth}{Last-25\%-steps alternating replay (token-balanced)} & pass@1 & \cellcolor{yellow!25}\textit{71.3} & \textit{30.5} & \cellcolor{yellow!25}\textbf{2.2} & \cellcolor{yellow!25}\textit{11.8} & 6.8 & 6.0 & \cellcolor{yellow!25}\underline{46.7} & 67.4 \\
 & pass@8 & \cellcolor{yellow!25}\textbf{88.6} & \cellcolor{yellow!25}\textbf{48.0} & \cellcolor{yellow!25}\textbf{11.5} & 24.6 & 17.7 & 17.8 & \cellcolor{yellow!25}\textbf{62.7} & 87.7 \\
 & pass@32 & \underline{93.2} & \cellcolor{yellow!25}\textbf{57.4} & \cellcolor{yellow!25}\underline{22.0} & \underline{30.0} & 26.7 & \textit{26.7} & \cellcolor{yellow!25}\textbf{69.5} & \textbf{94.1} \\
\midrule[1.5pt]
\multicolumn{10}{c}{Last-50\%-steps}\\
\midrule
\multirow{3}{0.22\linewidth}{Last-50\%-steps alternating replay} & pass@1 & \cellcolor{yellow!25}71.1 & 30.3 & \cellcolor{yellow!25}\textit{1.7} & \cellcolor{yellow!25}\textbf{12.5} & 6.8 & 6.0 & \cellcolor{yellow!25}\textit{46.5} & \textbf{68.8} \\
 & pass@8 & \textit{88.4} & \textit{47.0} & \cellcolor{yellow!25}\underline{10.0} & \cellcolor{yellow!25}\textbf{26.8} & \textit{19.4} & 17.2 & 62.3 & \textbf{88.7} \\
 & pass@32 & 92.6 & \cellcolor{yellow!25}\underline{56.6} & \cellcolor{yellow!25}\textbf{23.0} & \cellcolor{yellow!25}\textbf{33.3} & \textit{30.0} & \cellcolor{yellow!25}\underline{30.0} & \cellcolor{yellow!25}\underline{69.4} & \underline{93.9} \\
\midrule
\multirow{3}{0.22\linewidth}{Last-50\%-steps alternating replay (token-balanced)} & pass@1 & \cellcolor{yellow!25}\textbf{71.5} & \cellcolor{yellow!25}\textbf{30.9} & 1.4 & \cellcolor{yellow!25}\textit{11.8} & \cellcolor{yellow!25}\textbf{8.8} & \cellcolor{yellow!25}\textbf{6.7} & \cellcolor{yellow!25}\textbf{46.9} & \underline{68.5} \\
 & pass@8 & 88.3 & \underline{47.9} & 8.1 & \textit{25.8} & \cellcolor{yellow!25}\textbf{25.0} & \cellcolor{yellow!25}\textit{18.8} & \underline{62.5} & \underline{88.4} \\
 & pass@32 & 92.4 & \cellcolor{yellow!25}\underline{56.6} & 17.0 & \underline{30.0} & \cellcolor{yellow!25}\textbf{36.7} & \cellcolor{yellow!25}\underline{30.0} & 68.8 & \textit{93.7} \\
\bottomrule
\end{tabular}%
}
\end{table}